\documentclass[iicol,pdflatex,sn-chicago]{sn-jnl}

\usepackage{amsmath,amssymb,amstext,amsthm, bm, mathtools}
\usepackage{graphicx}
\usepackage{subcaption}
\usepackage{float}
\usepackage{duckuments}

\usepackage{layouts}

\usepackage[linesnumbered,ruled,vlined]{algorithm2e}
\usepackage{cuted}

\PassOptionsToPackage{hyphens}{url}\usepackage{hyperref}
\usepackage{cleveref}
\usepackage{booktabs}
\usepackage{siunitx}
\usepackage{tabularx}

\DeclareMathOperator*{\argmin}{argmin}

\newtheorem{theorem}{Theorem}[section]
\newtheorem{lemma}[theorem]{Lemma}

\SetKwInOut{KwInput}{Input}
\SetKwInOut{KwOutput}{Output}

\SetCommentSty{commfont}

\SetFuncSty{funcfont}

\newcommand{\rememberlines}{\xdef\rememberedlines{\number\value{AlgoLine}}}

\usepackage{xcolor}

\newcommand{\edit}[1]{{#1}}
\newcommand{\finaledit}[1]{{#1}}

\newcommand*{\mA}{\ensuremath{\mathcal{A}}}
\newcommand*{\mC}{\ensuremath{\mathcal{C}}}
\newcommand*{\mE}{\ensuremath{\mathcal{E}}}
\newcommand*{\mG}{\ensuremath{\mathcal{G}}}
\newcommand*{\mI}{\ensuremath{\mathcal{I}}}
\newcommand*{\mT}{\ensuremath{\mathcal{T}}}
\newcommand*{\mX}{\ensuremath{\mathcal{X}}}
\newcommand*{\mV}{\ensuremath{\mathcal{V}}}

\DeclareMathOperator{\pr}{\text{Pr}}  %
\DeclareMathOperator{\var}{\text{VaR}}
\DeclareMathOperator{\cvar}{\text{CVaR}}

\renewcommand{\orcidlogo}{%
  \includegraphics[width=10pt]{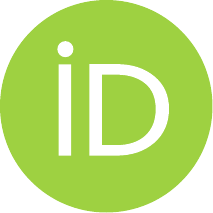}%
}
\renewcommand{\orcid}[1]{\href{#1}{\orcidlogo}}

\makeatletter
\newcommand\thefontsize{The current font size is: \f@size pt}
\makeatother

\makeatletter
\let\orig@maketitle\@maketitle
\renewcommand{\@maketitle}{\vspace{-41pt}\orig@maketitle}
\makeatother

\begin{document}

\title{Risk-Averse Traversal of Graphs with Stochastic and Correlated Edge Costs for Safe Global Planetary Mobility}

\author*[1]{\fnm{Olivier} \sur{Lamarre} \orcid{https://orcid.org/0000-0002-2249-7658}} \email{olivier.lamarre@robotics.utias.utoronto.ca}

\author[1]{\fnm{Jonathan} \sur{Kelly} \orcid{https://orcid.org/0000-0002-5528-6136}} \email{jonathan.kelly@robotics.utias.utoronto.ca}

\affil*[1]{\orgdiv{Space \& Terrestrial Autonomous Robotic Systems (STARS) Laboratory}, \orgname{University of Toronto Institute for Aerospace Studies}, \orgaddress{\street{4925 Dufferin Sreet}, \city{Toronto}, \postcode{M3H 5T6}, \state{Ontario}, \country{Canada}}\vspace{-0.5cm}}

\abstract{
Strategic mobility planning for robotic planetary surface exploration
is an important task that involves finding candidate long-distance routes on orbital maps and identifying segments with uncertain traversability.
Then, expert human operators establish safe, adaptive traverse plans based on the actual navigation difficulties encountered in these uncertain areas.
In this paper, we formalize this task as a new, risk-averse variant of the Canadian Traveller Problem (CTP) tailored to global planetary mobility.
The objective is to find a traverse policy minimizing a conditional value-at-risk (CVaR) criterion, which is a risk measure with an intuitive interpretation.
We propose a novel search algorithm that finds exact CVaR-optimal policies.
Our approach leverages well-established optimal AND-OR search techniques intended for (risk-agnostic) expectation minimization and extends these methods to the risk-averse domain.
We validate our approach through simulated long-distance planetary surface traverses; we employ real orbital maps of the Martian surface to construct problem instances and use terrain maps to express traversal probabilities in uncertain regions.
Our results illustrate different adaptive decision-making schemes depending on the level of risk aversion.
Additionally, our formulation accommodates traversability correlations between similar regions of the environment.
In such cases, we empirically demonstrate how information-seeking detours can effectively mitigate risk.
}

\keywords{Risk awareness, planetary mobility, path planning, autonomy}
\maketitle

\section{Introduction}
\label{intro}

Most planetary rovers drive several kilometres or more to maximize the science return of their missions.
An important operational component of long-range mobility is strategic planning, which consists of inspecting orbital imagery and finding candidate global routes connecting regions of scientific interest.
Subsequently, these strategic traverse plans guide the local, short-range driving efforts taking place on a regular basis~\citep{verma_first_2022}.
During surface exploration missions, rovers constantly travel through partially unknown regions of the environment.
Adaptive navigation planning, which enables global route adjustments, is necessary.

\begin{figure}
    \centering
    \includegraphics[width=\linewidth]{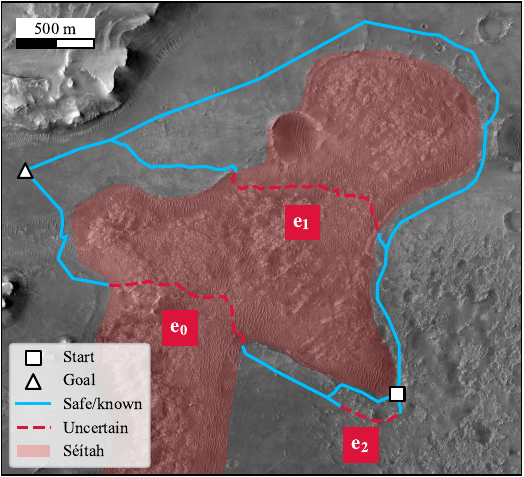}
    \caption{
        Conceptual example of the problem addressed in this paper. %
        A rover on the Jezero Crater floor, Mars, must reach a designated goal location by following the global route network shown.
        This graph has edges cutting through S\'e\'itah, a region with uncertain traversability.
        The rover may drive to the start of these edges to observe their traversability or follow a safe (but physically longer) route around S\'e\'itah. %
        Here, prioritizing the shortest traverse through $e_0$ may be costly in the worst case; if this uncertain edge happens to be untraversable (or traversable at a very high cost), the rover would need to backtrack and follow a different route.
        A risk-averse strategy might forego $e_0$ and head towards $e_1$ or stick to the (known) blue edges only from the start.
        Accounting for traversability correlations between edges (e.g., based on underlyling terrain properties) would enable different risk aversion behaviours.
        For instance, if the terrain along $e_0$ and $e_2$ is very similar, a risk-averse strategy may prioritize a short information-seeking drive to the latter before deciding whether it is worth heading towards $e_0$.
        }
    \label{fig:overview}
\end{figure}

Human operators are still heavily involved in mobility planning for current rover missions, in part because of their ability to learn from past experiences and devise safe traverse strategies~\citep{lamarre_importance_2024}.
For example, the wheel degradation issue on board the Curiosity rover was, in part, mitigated by small detours through less risky terrain.
Global geomorphic maps were incrementally updated with observations collected on the Martian surface to keep human operators informed about the possible location of wheel-damaging sharp rocks~\citep{arvidson_relating_2017}.
Additionally, at the beginning of the Mars 2020 mission, a long detour around a precarious ripple field was preferred over a physically shorter traverse cutting through the field.
Instead of risking wasting precious time contending with mobility challenges along the shorter route, human operators entrusted the rover's autonomous navigation software to complete the long traverse quickly and reliably~\citep{rankin_perseverance_2023}.
This confidence in the robot's autonomy came from prior successful test drives on terrain similar to that along the longer route.
\edit{Later in the mission, a higher-than-expected density of boulders slowed the traverse through the Margin Unit, a region near the Jezero Crater rim, by forcing short drives manually guided by human experts.
Mission operators eventually deviated from the original strategic route and entered a nearby ancient river valley, where they knew the simpler terrain would allow long and fast autonomous drives again~\citep{noauthor_nasas_2024}.}

In this paper, we propose a framework combining risk-averse and adaptive behaviours tailored to global route planning on planetary surfaces.
We represent the network of candidate traverses as a graph containing edges with correlated random costs (called \textit{stochastic} edges). %
As the rover drives towards a given goal vertex, it observes the actual (true) cost of stochastic edges and dynamically refines its belief over the cost of remaining, uncertain edges.
Our method finds safe graph traversal policies that minimize a conditional value-at-risk (CVaR) criterion, which is a risk measure with an intuitive interpretation.
We conceptually illustrate the importance of risk aversion in this context in~\Cref{fig:overview}.
The contributions of this work are threefold:
\begin{enumerate}
    \item Formulation of a novel risk-averse variant of the Canadian Traveller Problem (CTP), termed CVaR-CTP, where the costs of stochastic edges may be correlated and the objective is to minimize a CVaR objective.
    \item %
    A forward search method that returns CVaR-optimal policy trees. Unlike existing approximate dynamic programming paradigms, our approach finds exact solutions. Efficiency mechanisms that do not compromise solution optimality are also presented.
    \item Realistic demonstrations of global traverses using real Martian orbital maps. We open-source\footnote{\finaledit{Available at \href{https://github.com/utiasSTARS/qgis-planetary-graph-tool}{https://github.com/utiasSTARS/qgis-planetary-graph-tool}}} a custom QGIS tool to draw strategic route networks and to extract terrain map information along edges.
\end{enumerate}

This paper is structured as follows.
\Cref{sec:relatedwork} discusses graph search applied to uncertain environments and the optimization of a CVaR criterion.
Next, the central problem of this paper is formally described in~\Cref{sec:problemstatement}.
Our proposed approach is then detailed in~\Cref{sec:approach}.
Lastly, experimental results are presented and discussed in~\Cref{sec:experiments}.

\section{Related Work}
\label{sec:relatedwork}

We begin by reviewing graph search problems applied to long-range mobility in partially unknown environments.
Then, we shift the discussion to the use of safe feedback control laws to solve such problems.
We limit our review to risk-sensitive policies and focus primarily on objective functions that involve the conditional value-at-risk (CVaR).

\subsection{Graph Search for Uncertain Environments}

One of the simplest planning strategies for an agent driving through an unknown environment is to assume a deterministic (known) cost model and react to unexpected cost observations.
For example, Lifelong Planning A* and D* Lite~\citep{koenig_fast_2005} are incremental search algorithms that only adjust their solutions when new traversability information becomes available.
In this work, we instead proactively account for the uncertainty induced by incomplete knowledge of the environment.

The Canadian Traveller Problem (CTP) \citep{papadimitriou_shortest_1991} consists of traversing a graph that contains edges that might be blocked.
An agent may disambiguate the status of such stochastic edges (i.e., determine whether they are traversable or not) by observing them from one of their incident vertices.
Papadimitriou and Yannakakis also describe a stochastic variant in which edge blockage probabilities are known a priori and are assumed to be independent.
In the current paper, `CTP' refers to the stochastic variant.
This problem is solved optimally with a policy that minimizes the expected total traverse cost between (given) start and goal vertices, which is \mbox{\#P-hard}~\citep[Theorem 4.4]{papadimitriou_shortest_1991}.

Many planning problems under uncertainty are extensions of the CTP.
The stochastic shortest path problems with recourse (SSPPR) describes two similar models: one in which probabilities over graph cost realizations are given, and another in which the cost of individual edges follow independent, discrete probability distributions~\citep{polychronopoulos_stochastic_1996}.
The latter problem is also \#P-hard~\citep[Theorem 4]{polychronopoulos_stochastic_1996}.
The authors propose dynamic programming algorithms to find exact solutions and heuristic methods as approximations for larger problem instances.
In~\citep{bnaya_canadian_2009}, the CTP is augmented with the possibility to disambiguate distant stochastic edges at a known cost.
The Bayesian CTP~\citep{lim_shortest_2017} assumes a probability distribution over the hypothesis space of possible graph realizations, similar to one of the SSPPR variants~\citep{polychronopoulos_stochastic_1996}.
The Partial Covering CTP looks for policies that guide an agent through a given target set of vertices~\citep{huang_stochastic_2023}.
The framework in~\citep{cao_anmip_2023} implements a rollout-based CTP solver and incrementally estimates independent edge blockage probabilities by repeatedly traversing the same environment.

As initially highlighted by~\cite{ferguson_pao_2004}, a CTP instance can be represented as an AND-OR (AO) graph and solved optimally using heuristic search methods such as AO*~\citep[Chapter 3]{nilsson_principles_1982}.
From a given start state, AO* grows a tree through the underlying graph.
This methodology only explores promising, reachable regions of the state space, often a benefit over other dynamic programming approaches that typically iterate over the entire space.
The authors of~\citep{ferguson_pao_2004} introduce PAO*, a variant of AO*, which reduces computational effort through improved information sharing between nodes of the search tree.
The CAO* algorithm incorporates additional efficiency mechanisms, such as node pruning and caching~\citep{aksakalli_ao_2016}.
The same authors also demonstrate how the CTP can be recast as a Markov Decision Process (MDP) or a partially observable MDP with deterministic observations (deterministic POMDP).

Most CTP problems seek policies with minimal expected cost; solutions are agnostic to the spread of cost outcomes.
The Robust CTP (RCTP)~\citep{guo_robust_2019} instead optimizes an exponential utility risk measure of the total cost incurred.
Surprisingly, optimal policies are found with minor modifications to AO*.
In~\citep{guo_dual_2022}, approximate policies that minimize a mean-standard deviation risk measure are found.
This work assumes known and independent edge blockage probabilities, which is common with most CTP-like problems.
In reality, as explored in the current paper, different regions of an uncertain field environment may have correlated traversability.

\subsection{Risk-Averse Decision-Making Using a CVaR Criterion}

Risk-averse optimization for sequential decision-making tasks has been studied for decades.
In the early 1970s, the authors of~\citep{howard_risk-sensitive_1972} found Markov policies using a dynamic programming formula that employed an exponential utility risk measure instead of the conventional (risk-neutral) expectation.
This substitution was the same one used in the aforementioned AO* algorithm variant for the RCTP.
Many other paradigms using a variety of risk measures have since been proposed for the safe control of autonomous systems~\citep{wang_risk-averse_2022}.

The conditional value-at-risk (CVaR), formally introduced in the next section, is a risk measure that stands out because of its intuitive interpretation: it is equivalent to an expectation over a fraction of worst outcomes.
In sequential decision-making problems, a CVaR-optimal policy therefore minimizes the expectation over a tail of its total (cumulative) cost distribution.
Finding such policies is computationally challenging because the CVaR is not time-consistent~\citep[Section 5]{majumdar_how_2020}.
Consequently, Bellman's principle of optimality is not satisfied, precluding the straightforward use of dynamic programming.
Additionally, CVaR policies are generally history-dependent (non-Markovian) because of the measure's dependence on the cumulative cost of a decision process.
\cite{bauerle_markov_2011} overcome this issue by augmenting the state space of a baseline MDP with the cumulative cost incurred by an agent.
They propose an approximate solution that relies on repetitively finding time-consistent policies with dynamic programming.
Another state space augmentation approach is proposed by~\cite{chow_risk-sensitive_2015}, while~\cite{rigter_risk-averse_2021} adopt the game-theoretical view of~\citep{chow_risk-sensitive_2015} and use a Monte Carlo tree search-based solution to find approximate policies.
Some methods circumvent the difficulties associated with dynamic programming for CVaR and instead optimize policies directly.
For example,~\cite{tamar_policy_2015} propose a gradient formula for the CVaR and use it in a risk-sentitive policy gradient scheme.
\cite{wang_adaptive_2021} employ stochastic search to iteratively optimize the parameters of a distribution from which policies are sampled and applied over a fixed horizon.

Other works involve the CVaR in decision-making problems beyond simply making it the central optimization criterion.
Approaches based on policy gradient and actor-critic are presented in~\citep{chow_algorithms_2014} to find risk-neutral solutions subject to an upper bound on their CVaR.
In~\citep{rigter_planning_2022}, a policy with the lowest expected cost is found among the CVaR-optimal solution set.
It is also possible to craft a \textit{dynamic} risk measure by calculating the CVaR at every stage of a process instead of computing the CVaR of the cumulative cost incurred by an agent (referred to as a \textit{static} risk measure).
This modified formulation maintains time consistency and can be solved with dynamic programs~\citep{ruszczynski_risk-averse_2010,majumdar_how_2020}.
It is unclear, however, what compounding many CVaR measures effectively amounts to as far as risk aversion is concerned; although mathematically convenient, dynamic risk measures generally lack the intuitive interpretation of static risk measures~\citep[Section 4.2.2]{wang_risk-averse_2022}.

Mobility frameworks for field robots have relied on the CVaR to find risk averse paths instead of policies.
One of the planners in~\citep{dixit_step_2024} minimizes a criterion combining path length and a dynamic CVaR measure.
In~\citep{endo_risk-aware_2023}, a deterministic planner uses costmaps generated with risk-averse rover slip estimates.
The EVORA pipeline~\citep{cai_evora_2024} implements two receding horizon path planners optimizing different CVaR criteria, one of which extends the stochastic search method presented by~\cite{wang_adaptive_2021}.
In the current paper, we instead focus on finding CVaR-optimal graph traversal policies.%
\section{Problem Statement}
\label{sec:problemstatement}

We propose a risk-averse variant of the CTP in which the objective is to find CVaR-optimal policies.
A brief mathematical definition of the CVaR measure is provided below and the central optimization problem of this paper is formulated.

\subsection{Conditional Value-at-Risk (CVaR)}

Let $Z$ represent a real-valued random cost with the cumulative \edit{distribution} function $F_Z(z)=\pr(Z \leq z)$.
Given parameter $\alpha \in (0,1)$, the value-at-risk (VaR) of $Z$ is a quantile function
\begin{equation}
    \var_\alpha(Z) = F_Z^{-1}(1-\alpha) = \inf \{z : F_Z(z) \geq 1-\alpha\}.
\end{equation}
The conditional value-at-risk (CVaR) is defined as the average\footnote{In finance and generic stochastic optimization, CVaR is sometimes called \textit{average value-at-risk} (AVaR) or \textit{expected shortfall}.} of the VaR over the interval $\left[\alpha,0\right)$:
\begin{equation} \nonumber
    \cvar_\alpha(Z) = \frac{1}{\alpha}\int_{1-\alpha}^{1} \var_{1-u}(Z)\,du.
\end{equation}
The CVaR can also be determined by solving an optimization problem:
\begin{equation} \label{eq:cvar-inf}
    \cvar_\alpha(Z) = \inf_{s \in \mathbb{R}} \Bigl\{ s + \frac{1}{\alpha}\mathbb{E}\bigl[(Z-s)^+\bigr]\Bigr\},
\end{equation}
where ${(\zeta)^+ = \text{max}(\zeta,0)}$ and as before, ${\alpha \in (0,1)}$ \citep[Section 3.1]{wang_risk-averse_2022}.
The minimizer is ${s^*=\var_\alpha(Z)}$ and it intuitively corresponds to the cost threshold above which $Z$ influences the CVaR calculation.
A smaller $\alpha$ results in a more risk-averse evaluation of the random cost.
At the extremes, $\alpha \rightarrow 0$ approaches the behaviour of a worst-cost risk measure, while $\alpha = 1$ is equivalent to the standard (risk-neutral) expectation.
In the event that $Z$ follows a discrete distribution with finite support $\Omega_Z$, the optimization problem in \Cref{eq:cvar-inf} becomes a combinatorial one:
\begin{equation} \label{eq:cvar-min}
    \cvar_\alpha(Z) = \min_{s \in \Omega_Z} \Bigl\{ s + \frac{1}{\alpha}\mathbb{E}\bigl[(Z-s)^+\bigr]\Bigr\}.
\end{equation}

\subsection{CVaR-CTP Formulation}
\label{sec:formulation}

Our problem is an extension of the standard CTP.
\edit{
Let $\mG=(\mV, \mE)$ be a graph with the vertex set $\mV$ and weighted edge set $\mE$, where the edges may be directed or undirected.
The graph represents a network of paths through a partially-known environment.
The start and goal vertices $v_0,v_g \in \mV$ are assumed to be known and the set of edges is split in two: subset $\mE_d$ includes traversable, deterministic edges (whose costs are known a priori), while subset $\mE_s$ includes stochastic edges (whose traversability statuses are unknown initially).
Let $m_s=\lvert \mE_s \rvert$ denote the number of stochastic edges and let $\mV_s \subset \mV$ represent the subset of vertices incident to those edges.
As in a typical CTP, an agent navigating through $\mG$ may take the shortest deterministic path (i.e., a path only containing deterministic or observed stochastic edges) to a reachable\footnote{By contrast, unreachable vertices are those that would require traversing an unobserved stochastic edge to get to.} vertex in $\mV_s$ and, from there, observe an adjacent stochastic edge to reveal its actual cost.
We assume that observed stochastic edges maintain their actual cost until the end of the traverse; we effectively treat them as deterministic.
Alternatively, the agent may take the shortest deterministic path to the goal vertex $v_g$ if such a path exists.

In the standard CTP, observing a stochastic edge yields one of two possible outcomes with different probabilities: the edge is either traversable at a positive finite cost, or it is blocked and effectively has infinite cost.}
In this work, we use a slightly more permissive formulation: stochastic edges can have either a low-cost status or a high-cost status.\footnote{In this paper, we only consider two possible stochastic edge traversability statuses. The algorithm presented in the next section, however, supports any finite number of edge cost realizations.}
The latter need not be infinite\edit{---this formulation accommodates problem instances in which stochastic edges are traversable with varying levels of difficulty or effort.
For example, as demonstrated with Perseverance in the Margin Unit, the cost of traversing a terrain region may depend on the level of navigation autonomy that can be employed.
Unobstructed terrain is conducive to autonomous driving (fast, low cost) while rugged terrain may cause autonomy failures and require manually-guided drives (slow, high cost, but still ultimately traversable).
}

\edit{
Similar to the (standard, risk-agnostic) CTP description in~\cite[Section 2.1]{aksakalli_ao_2016}, our problem can be cast in a risk-averse finite Markov decision process (MDP) framework.
The following text describes the corresponding state space, action space, cost function, and state transition probabilities.}
At any given time, the complete state of an agent is characterized by \edit{the agent's location (a vertex) in the graph, and its knowledge of the traversability status of all $m_s$ stochastic edges.
The former is an element of the set ${\mathcal{Y} \coloneqq \{v_0,v_g\} \cup \mV_s}$, which is the set of vertices from which the agent takes actions or terminates its traverse.
The latter is an element from the space of information sets ${\{\text{L},\text{H},\text{A}\}^{m_s}}$, where the labels L, H and A correspond to a low-cost status, a high-cost status, and an ambiguous status (not yet observed), respectively.
An element $\mI \in {\{\text{L},\text{H},\text{A}\}^{m_s}}$ is an ordered set of cardinality $m_s$ that indicates the knowledge of all $m_s$ stochastic edges.
As such, we define the overall state space as $\mX \coloneqq \mathcal{Y} \times \{\text{L},\text{H},\text{A}\}^{m_s}$.
Since none of the stochastic edge statuses are known at the start, the initial state $x_0=(v_0,\,\mI_0)$ has the information set ${\mI_0=\{\text{A},\text{A},...,\text{A}\}}$.
From an arbitrary state $x=(v,\mI) \in \mX$, the following set of actions $\mA(x)$ is available to the agent:
\begin{itemize}
    \item Drive-observe action: take the shortest deterministic path from $v$ to a reachable vertex $v' \in \mV_s$ incident to an unobserved stochastic edge and observe this edge;
    \item Drive-terminate action: take the shortest deterministic path to the goal vertex $v_g$ if such a path exists.
    This action ends a traverse since the resulting state is terminal.
\end{itemize}

The cost of an action, captured by the function ${c:\mX \times \mA \rightarrow \mathbb{R}_{\geq 0}}$, is the sum of individual edge costs along the path taken, which we assume is always non-negative\footnote{The cost of a single drive-observe action may be 0 if it consists of staying at the current vertex to observe another adjacent stochastic edge.} and finite.
The costs of deterministic edges are known from the start while the cost of observed stochastic edges depend upon their traversability status (L or H), as encoded in the information set $\mI$ of the state from which the action is taken.
For feasibility reasons, we assume that if all stochastic edges have a high-cost status (rendering them untraversable), a drive-terminate action from the start vertex always has a finite (but possibly very large) cost.
In the event that an infinite cost is assigned to all high-cost stochastic edges, this assumption ensures that a path composed solely of deterministic edges connects the start and goal vertices.

A state transition probability function $\tau:\mX \times \mA \times \mX \rightarrow [0,1]$ quantifies the probabilities of action outcomes.
Since the agent only traverses paths with known finite costs regardless of the action $a \in \mA$ taken, the transition from the originating vertex $v$ to the destination vertex, represented with $v'=\text{dest}(a)$, is always deterministic.
If action $a$ is a drive-observe action, the outcome of the post-drive observation (whether the observed stochastic edge, represented with $e=\text{edge}(a)$, has a low-cost or high-cost status) is random.
In the standard CTP, this stochastic part of the state transition dynamics is represented with a single function $p: \mE_s \rightarrow [0,1]$, which assigns every stochastic edge a high-cost status probability that remains constant throughout the traverse.
However, in planetary exploration, it is far more common to dynamically update traversability knowledge based on the information gathered.
As such, we define a more generic map ${\rho: \mE_s \times \{\text{L},\text{H},\text{A}\}^{m_s}  \rightarrow [0,1]}$, which is a user-defined, problem-specific probability function conditioned on the current information set, akin to a Bayes-Adaptive MDP~\citep{duff_design_2003}.
For example, if multiple candidate probability functions are under consideration ($p_0, p_1, ...$), then $\rho$ may be a belief over this set of functions that is refined as more observations are gathered.
The transition from current state ${x=(v,\mI)}$ to a successor state ${x'=(\text{dest}(a),\mI')}$ when taking a drive-observe action ${a \in \mA(x)}$ occurs with probability
\begin{multline}  \label{eq:tau}
    \tau((\text{dest}(a),\mI') \mid (v,\mI), a) =\\*[1ex]
    \begin{cases}
        \rho(\text{edge}(a) \mid \mI) &
        \begin{aligned}
            &\text{if the observed edge} \\
            &\text{has status H in $\mI'$},
        \end{aligned}\\[3ex]
        1 - \rho(\text{edge}(a) \mid \mI) &
        \begin{aligned}
            &\text{if the observed edge} \\
            &\text{has status L in $\mI'$},
        \end{aligned}\\[3ex]
        0 &
        \begin{aligned}
            &\text{for all other (invalid)} \\
            &\text{information sets $\mI'$}.
        \end{aligned}
    \end{cases}
\end{multline}
The successor information set $\mI'$ is defined to be the same as $\mI$ except for the (updated) status label of the observed edge.
We note that the conditioning of $\rho$ on the information set is optional; our formulation is also compatible with functions outputting constant probabilities as in the standard CTP.
}

\edit{
We define an instance of the CVaR Canadian Traveler Problem (CVaR-CTP) as a tuple
\[
\left<\mX, \mA, v_0, v_g, c, \tau, \alpha\right>
\]
and it is solved with a policy that guides an agent from start vertex $v_0$ to goal vertex $v_g$.
Here, the underlying graph $\mG=(\mV,\mE)$ is represented by the state and action spaces ($\mX$ and $\mA$).
In this context, an optimal policy $\pi^*$ minimizes a CVaR criterion: the total traverse cost at risk aversion level $\alpha$ from the start state ${x_0 = (v_0,\{\text{A,A,...A}\})}$.
The optimization problem is:
\begin{align} \label{eq:opt}
    J_{\pi^*}(x_0) &\mathbin{=} \min_{\pi \in \Pi} \; \cvar_\alpha (\mathrm{C}^\pi(x_0)) \nonumber \\
    &\mathbin{=} \min_{\pi \in \Pi} \; \min_{s \in \mC^\pi(x_0)} \Bigl\{s\!+\!\frac{1}{\alpha}\mathbb{E}\bigl[(\mathrm{C}^\pi(x_0)-s)^+\bigr] \Bigr\},
\end{align}
where $\Pi$ is the set of all policies and the total cost of some policy $\pi$ from start state $x_0$, $\mathrm{C}^\pi(x_0)$, follows a discrete probability distribution $\mC^\pi(x_0)$.
This distribution has finite support because there is at most a finite number of stochastic edges to observe and each observation is associated with a finite number of outcomes (two possible traversability statuses in this case).
Thus, the substitution of CVaR in the main optimization formula follows \Cref{eq:cvar-min}.
As mentioned in \Cref{sec:relatedwork}, a CVaR-optimal policy depends on the process trajectory and is therefore non-Markovian.
In the next section, however, we introduce a forward search algorithm that circumvents this difficulty by representing policies as decision trees, which are data structures that inherently encode the process history.
}

\section{CVaR-Optimal AND-OR Search}
\label{sec:approach}

We propose an optimal offline algorithm to solve CVaR-CTPs.
In short, our approach grows an AO tree through \edit{the state space starting at the given initial state $x_0$.}
Optimizing a risk-averse criterion does not preclude us from leveraging computational efficiency mechanisms used by previous risk-neutral AO* variants.
Additionally, in the event that several CVaR-optimal policies exist, our implementation retains the one with the lowest expected cost.

We begin with an overview of the classical AO* algorithm and demonstrate it on a simple CTP instance.
Then, we detail our proposed approach for solving CVaR-CTPs.

\subsection{AND-OR Search for Graph Traversal Under Uncertainty}

Our approach builds upon the work discussed in \Cref{sec:relatedwork}, in which CTPs are expressed as AO trees~\citep{ferguson_pao_2004, aksakalli_ao_2016, guo_robust_2019, huang_stochastic_2023}.
An AO tree is composed of OR nodes and AND nodes connected by directed arcs.
A node corresponds to a state in the underlying CTP and an arc corresponds to a state transition.
Successors of OR nodes are AND nodes and vice versa.

The root of the tree is an OR node representing the starting state $x_0=(v_0,\mI_0)$.
OR nodes represent states from which actions are taken.
Each action is depicted with an outgoing deterministic arc connecting to an AND node.
For a drive-terminate action, the AND node is terminal (i.e., does not have outgoing arcs) and corresponds to a state at the goal vertex $v_g$.
For a drive-observe action, the AND node corresponds to the rover state at a vertex incident to a stochastic edge with this edge still labelled as ambiguous/unobserved (A).
The possible edge traversability statuses (in this case, low-cost (L) or high-cost (H)) are each represented by arcs connecting the AND node to its successor OR nodes.
These new OR nodes correspond to the new state following the corresponding observation and the incoming arcs are associated with the observation probability retrieved from the underlying CTP.
\edit{A ``frontier'' node does not have successors, either because it has not yet been expanded or because it corresponds to a terminal state (a leaf node).}

Since the number of nodes in the tree increases exponentially with depth (where the depth corresponds to the number of stochastic edges), explicit representations of the complete AO tree is limited to very small CTP instances.
The AO* algorithm instead incrementally grows an AO tree in promising regions of the search space until a complete policy (a subset of the main search tree) is found~\citep[Chapter 3]{nilsson_principles_1982},~\citep{martelli_additive_1973}.
AO* search iterations are broken down into three steps:
\begin{enumerate}
    \item Node selection: a non-terminal frontier node of the best partial policy\footnote{A partial policy is a subtree containing some frontier (unexpanded) nodes that do not correspond to terminal states. In contrast, a complete policy is a full decision-making scheme in which all frontier nodes correspond to terminal states (all possible outcomes of that policy).} found so far is selected.
    \item Node expansion: the selected node is expanded into its successor nodes and their costs are initialized with the help of an admissible heuristic function. Nodes corresponding to terminal states (i.e., rover at the goal vertex of the CTP) are marked as solved.
    \item Backpropagation: the cost estimates of the newly-created nodes are propagated through their ancestors up the search tree. The best partial policy tree is simultaneously updated. Additionally, OR nodes are marked as solved if their best successor is also solved, and AND nodes are marked as solved if all their successor nodes are solved.
\end{enumerate}
The AO* algorithm terminates when the root OR node is marked as solved.
Similar to the A* algorithm, the use of an admissible heuristic function ensures that AO* always terminates with an optimal solution~\citep{martelli_additive_1973}.
When applied to the CTP, an optimal policy has a minimum expected cost~\cite[Section 3.2]{aksakalli_ao_2016}.

\begin{figure}
    \centering
    \includegraphics[width=\columnwidth]{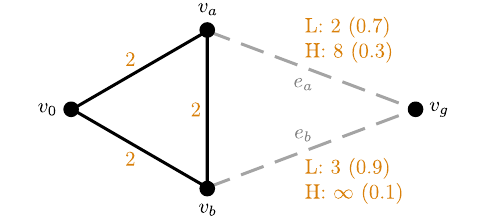}
    \caption{Sample CTP instance.
    An agent starts at vertex $v_0$ and must reach $v_g$.
    Solid edges are deterministic and have the indicated cost.
    Dashed edges are stochastic and their low-cost (L) and high-cost (H) statuses are specified.
    Status probabilities are independent and indicated in parentheses.
    }
    \label{fig:example-graph}
\end{figure}

\begin{figure*}
    \centering
    \includegraphics[width=\linewidth]{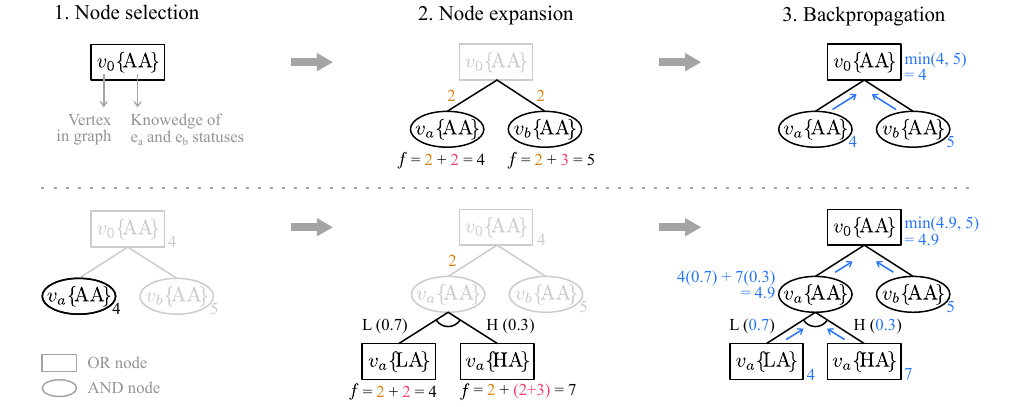}
    \caption{Node selection, node expansion and backpropagation steps of the first two AO* iterations applied to the CTP instance depicted in~\Cref{fig:example-graph}.
    OR and AND nodes are labelled with the corresponding state in the CTP instance. %
    In the node expansion step, the cost estimates of newly-expanded nodes ($f$) are the sum of a cost-to-come (orange) and a heuristic cost-to-go (pink).
    The latter is obtained through the assumption that unobserved stochastic edges have the lowest cost possible, i.e., a low-cost status (L).
    }
    \label{fig:example-aostar}
\end{figure*}

\begin{figure}
    \centering
    \includegraphics[width=\linewidth]{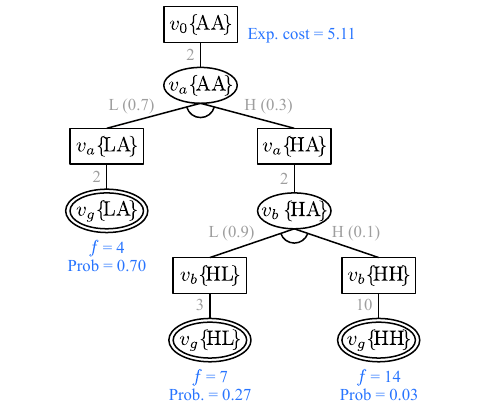}
    \caption{Optimal policy tree found with AO* applied to the CTP instance depicted in~\Cref{fig:example-graph}.
    The meaning of symbols and labels are the same as in~\Cref{fig:example-aostar}.
    The costs and probabilities at terminal AND nodes (doubly circled), which constitute the probability mass function of this policy, are included.
    }
    \label{fig:example-policy}
\end{figure}

\Cref{fig:example-graph} shows a sample CTP instance with two stochastic edges, while~\Cref{fig:example-aostar} shows the first two AO* iterations applied to this example.
An optimal policy tree (with the minimal expected cost) is shown in~\Cref{fig:example-policy}.
In this solution, the agent first moves to vertex $v_a$ to disambiguate stochastic edge $e_a$.
If a low-cost status (L) is observed, the agent heads straight to the goal vertex $v_g$.
Otherwise, the agent moves to vertex $v_b$ and disambiguates edge $e_b$.
If a low-cost status (L) is observed, the agent goes to $v_g$ via edge $e_b$.
Otherwise, the agent returns to $v_a$ and continues to $v_g$ via edge $e_a$.

\edit{
The AO* implementation depicted in \Cref{fig:example-aostar}  differs slightly from standard algorithm given in~\cite{martelli_additive_1973,aksakalli_ao_2016,guo_robust_2019} and elsewhere.
We initialize the frontier node costs with the sum of the cost-to-come and an admissible cost-to-go heuristic.
In contrast, the standard AO* algorithm initializes these costs using the cost-to-go heuristic only and accumulates stage-wise costs during backpropagation at OR nodes.
Although the mechanics differ, both approaches yield the same root node cost estimates and best partial policies at each iteration, making our implementation functionally equivalent.
In our formulation, the cost of a node is therefore the combination of costs incurred from the root/start node and expected remaining costs.
As CVaR-optimal policies depend on the process cost history, this setup provides the foundation for the algorithm detailed in the next section.
}

\subsection{CVaR-CTP-AO*}

\edit{
We now describe an AO*-inspired algorithm to solve CVaR-CTP instances with policy trees.
With minor modifications to the procedure in~\cite[Section 3]{bauerle_markov_2011}, the optimization problem in \Cref{eq:opt} is broken down into two nested subproblems:
\begin{equation} \label{eq:opt_mod}
    J_{\pi^*}(x_0) =  \min_{s \in \mC^{\Pi(x_0)}} \Bigl\{ s + \frac{1}{\alpha}w(x_0, s) \Bigr\},
\end{equation}
where $\mC^{\Pi(x_0)}$ represents the set of possible total costs incurred with any\footnote{The set $\mC^{\Pi(x_0)}$ is the set of all unique cumulative costs under any process trajectory (from start state $x_0$ to a terminal state) of any complete policy tree rooted at $x_0$. In other words, it is the set of all cumulative costs at leaf nodes of a completely-expanded search tree of a problem instance.} complete policy tree rooted at $x_0$, and an inner problem defined as
\begin{equation} \label{eq:inner_exp}
    w(x_0, s) \coloneqq \min_{\pi \in \Pi(x_0)} \mathbb{E}\bigl[(\mathrm{C}^\pi-s)^+\bigr]
\end{equation}
where $\Pi(x_0)$ denotes the set of all complete policy trees rooted at start state $x_0$ and $\mathrm{C}^\pi$ is the (random) total cost of policy tree $\pi$.
For the remainder of the paper, we represent sets $\Pi(x_0)$ and $\mC^{\Pi(x_0)}$ with $\Pi$ and $\mC^\Pi$, respectively, since it is implied that we work with policy \textit{trees} rooted at the start state.

Put simply, the inner optimization problem $w(x_0,s)$ computes an expectation over the following random variable: the total cost incurred under some optimal complete policy tree rooted at $x_0$, shifted by some value $s$ and truncated under the nonlinear $(\cdot)^+$ operator.
The nonlinearity prevents the expression $(\mathrm{C}^\pi-s)^+$ from being additively separable over time steps (as in a standard MDP) and creates a dependence over the history of the process.
Consider the hypothetical situation in which we are given an explicit, fully-expanded search tree $\mT$ of some CVaR-CTP instance and every node $n \in \mT$ represents a state $x \in \mX$.
To evaluate $w(x_0,s)$ (and find the corresponding optimal policy) for some given shift value $s$, one must initialize the cost-to-come at all leaf nodes (corresponding to terminal states) by summing up all costs incurred from the start node and evaluate the nonlinear expression.
Then, a backward induction technique (similar to the one shown in the AO* example in the previous section) propagates these results up the tree.
The overall process is summarized as follows:
\sbox0{$\sum$}  %
\sbox1{$\min$}  %
\begin{multline} \label{eq:induction}
    w(n,s) = \\*
    \begin{cases}
        (g(n)-s)^+ &
        \begin{aligned}
            &\text{if $n$ is a frontier} \\
            &\text{node},
        \end{aligned}\\[3ex]
        \min\limits_{\mathrlap{\hspace*{-0.5\wd1}{n' \in succ(n)}}} \;\; w(n',s) &
        \begin{aligned}
            &\text{if $n$ is a non-} \\
            &\text{terminal OR node},
        \end{aligned}\\[3ex]
        \sum\limits_{\mathrlap{\hspace*{-0.5\wd0}{n' \in succ(n)}}} prob(n,n')\cdot w(n',s) &
        \begin{aligned}
            &\text{if $n$ is a non-} \\
            &\text{terminal AND node}.
        \end{aligned}\\[3ex]
    \end{cases}
\end{multline}
Function $g(n)$ is the total cost incurred along some process trajectory up to frontier node $n$, $succ(n)$ is the set of successors of node $n$ in the search tree, and $prob(n,n')$ is the probability of transitioning from AND node $n$ to OR node $n'$ (corresponding to a stochastic edge observation probability, in this case).
This process solves the inner optimization problem in \Cref{eq:inner_exp} when the induction reaches root node $n_0$ and $w(n_0,s)$ is calculated.
In practice, we do not have access to the fully-expanded search tree of a CVaR-CTP instance; we introduce an iterative forward search approach next.

Our proposed algorithm, called CVaR-CTP-AO*, relies on an expansion-backpropagation search mechanism.
Similar to AO*, it leverages an admissible heuristic function and incrementally grows a search tree $\mT$ until a complete CVaR-optimal policy subtree is found.
After expanding a non-terminal frontier node, we solve the inner optimization problem in \Cref{eq:inner_exp} using the induction procedure in \Cref{eq:induction} for different candidate shift values $s \in \mC^\Pi$, resulting in multiple partial policy candidates.
Before continuing onto the next search iteration, CVaR-CTP-AO* compares these candidates partial solutions and retains the one with the lowest CVaR according to \Cref{eq:opt_mod}.
This AO tree-based approach provides benefits over dynamic programming paradigms that require iterating over the entire state space multiple times (e.g., value iteration):
\begin{enumerate}
    \item A search tree is a hierarchical data structure that encodes the history of a process.
    As such, the cost-to-come $g(n)$ at some frontier node can be retrieved by tracing its ancestry up to the root node.
    Consequently, the policies found (subtrees of the main search tree) are inherently defined over the set of process histories.
    \item All possible shift values $s \in \mC^\Pi$ are the total cost estimates at frontier nodes in the search tree.
    Since the number of frontier nodes grows during search, the set $\mC^\Pi$ is dynamically updated and does not need to be calculated ahead of time.
    The values stored in this set are exact, leading to exact solutions.
    Interpolation schemes, such as those needed with the augmented state spaces in~\cite{chow_risk-sensitive_2015} and~\cite{bauerle_markov_2011} (with continuous costs), are unnecessary.
    \item By design, unpromising regions of the state space, or those unreachable from the start state, are avoided by the forward search mechanism.
\end{enumerate}
}

\begin{algorithm}[!t]
\fontsize{8}{10.5}\selectfont  %
\caption{CVaR-CTP-AO*}\label{alg:cvar-aostar-main}
\DontPrintSemicolon

\SetKwProg{fn}{function}{:}{}
\SetKwProg{pr}{procedure}{:}{}
\SetKwFunction{FSN}{SelectNode}
\SetKwFunction{FEN}{ExpandNode}
\SetKwFunction{FISC}{InitializeSuccessorCosts}
\SetKwFunction{FFTB}{BackpropTree}
\SetKwFunction{FAB}{BackpropAncestors}
\SetKwFunction{FOrB}{BackpropOr}
\SetKwFunction{FAndB}{BackpropAnd}
\SetKwFunction{FBC}{BestS}

\SetKwData{T}{\finaledit{$\mT$}}
\SetKwData{Troot}{$root(\mT)$}
\KwInput{$\mG$, $v_0$, $\mI_0$, $v_g$, \edit{$c$, $\tau$,} $\alpha$}
\KwOutput{CVaR-optimal policy tree}
\BlankLine

\finaledit{Declare \T \tcp{Search tree initialization}}
\Troot $\gets$ OrNode($v_0$, $\mI_0$)

$s^* \gets h(\Troot)$  \tcp{Best (and only) shift value}
$\mC^\Pi \gets \{s^*\}$ \tcp{Set of candidate shift values}
$\pi^* \gets$ Policy(\Troot, $s^*$)

\BlankLine

\While{$\neg$solved(\Troot, $s^*$)}{

    $n \gets$ \FSN{\Troot, $\pi^*$} \label{alg:cvar-aostar-main:fsn}

    \BlankLine
    \finaledit{\tcp{Expand node into its successors, append them to the tree, initialize their costs}
    \T $\gets$ \T $\cup$ \FEN{\T, $n$, $\mG$, $c$, $\tau$, $v_g$}} \label{alg:cvar-aostar-main:fen}

    $\mC^\Pi_\text{new} \gets$ \FISC{\finaledit{\T}, $n$, $\mC^\Pi$}

    \BlankLine
    \finaledit{\tcp{Compute \Cref{eq:inner_exp} for all nodes and shift values}}
    \FFTB{\T, $\mC^\Pi_\text{new}$} \label{alg:cvar-aostar-main:fftb}

    \FAB{$n$, $\mC^\Pi$} \label{alg:cvar-aostar-main:fab}

    \BlankLine

    $\mC^\Pi \gets \mC^\Pi \cup \mC^\Pi_\text{new}$

    \finaledit{\tcp{Shift value minimizing \Cref{eq:opt_mod}}}
    $s^* \gets$ \FBC{\Troot, $\mC^\Pi$, $\alpha$} \label{alg:cvar-aostar-main:fbc}

    $\pi^* \gets$ Policy(\Troot, $s^*$)
}
\Return $\pi^*$

\BlankLine
\fn{\FISC{\finaledit{\T}, $n$, $S$}}{
    $S_\text{new} \gets \{\}$

    \For{$n' \in \text{succ}(n)$}{
        $f(n') \gets g(n', \T) + h(n')$ \label{alg:cvar-aostar-main:init}

        \lIf{$f(n') \notin S$}{$S_\text{new}.add(f(n'))$}

        \lForEach{$s' \in S$}{$w(n', s') \gets \max(f(n')-s', 0)$}
    }

    \Return $S_\text{new}$
}
\BlankLine
\pr{\FOrB{$n$, $S$}}{
    \ForEach{$s' \in S$}{
        $n_\text{best}' \gets \argmin_{n' \in \text{succ}(n)} w(n', s')$

        $w(n, s') \gets w(n_\text{best}', s')$

        $solved(n, s') \gets solved(n_\text{best}', s')$
    }
}

\BlankLine
\pr{\FAndB{$n$, $S$}}{
    \ForEach{$s' \in S$}{
        \finaledit{\tcp{Compute expectation}}
        $w(n, s') \gets \sum_{n' \in \text{succ}(n)} prob(n, n') \cdot w(n', s')$

        $solved(n,s') \gets \bigwedge_{n' \in \text{succ}(n)} solved(n', s')$
    }
}

\BlankLine
\pr{\FFTB{\T, $S$}}{
    \ForEach{$n \in$ SortedNodes(\T)}{ \label{alg:cvar-aostar-main:FFTB-sort}
        \uIf{$\text{succ}(n) = \emptyset$}{
            \ForEach{$s' \in S$}{$w(n, s') \gets \max(f(n)-s', 0)$}
        }
        \lElseIf{type($n$)=OR}{\FOrB{n, S}}
        \lElse{\FAndB{n, S}}
    }
}

\BlankLine
\pr{\FAB{$n$, $S$}}{
    $q \gets \{n\}$ \quad \tcp{FIFO queue}

    \While{$\neg empty(q)$}{
        $n \gets q.\text{pop}()$

        \lIf{type($n$)=OR}{\FOrB{n, S}}
        \lElse{\FAndB{n, S}}

        \lIf{$\text{parent}(n) \neq \emptyset$}{$q.\text{push}(\text{parent}(n))$}
        }
}

\BlankLine
\fn{\FBC{$n_0$, $S$, $\alpha$}}{
    \Return $\argmin_{s' \in S} \; s' + \alpha^{-1}\cdot w(n_0, s')$
}
\rememberlines
\end{algorithm}

Our approach is detailed in \Cref{alg:cvar-aostar-main}.
The AO tree's root OR node corresponds to the \edit{initial state $x_0=(v_0,\mI_0)$}.
Since this is a frontier node, a total cost estimate to the goal vertex is computed with an admissible heuristic\footnote{In this work, this admissible heuristic function is the smallest possible path cost to the goal vertex, assuming all yet unobserved stochastic edges have a low-cost traversability status (L).} function $h$ and stored as the first (and initially only) element in the set of all possible total traverse costs \edit{$\mC^\Pi$}.
At the beginning, this value \edit{is trivially defined as the optimal shift $s^*$.
During search, $s^*$ represents the running minimizer of \Cref{eq:opt_mod}.
The corresponding best partial policy, $\pi^*$, minimizes $w(n_0,s^*)$ as detailed in \Cref{eq:inner_exp}.}
Similar to AO*, the core of the algorithm consists of three steps: node selection, node expansion, and cost backpropagation.
At first, a non-terminal frontier node of the best partial policy tree is expanded into its successor nodes according to the CVaR-CTP dynamics (\Cref{alg:cvar-aostar-main:fen}).
\edit{The total cost estimates assigned to the newly-created nodes, denoted with $f$, is the sum of their cost-to-come $g$ (cost of all actions taken from the start node of the corresponding process trajectory, easily obtained by a search tree lookup) and a their admissible heuristic cost-to-go, which is again calculated with function $h$.}
For reasons explained next, we identify which of these $f$-values are not yet included in \edit{$\mC^\Pi$ and temporarily store them in set $\mC^\Pi_{\text{new}}$.
The functions ${w(n,s) \; \forall s \in \mC^\Pi}$ at these frontier nodes are evaluated in preparation of backpropagation.
}

\edit{The backpropagation step solves the inner optimization problem (\Cref{eq:inner_exp}) for all shift values $s$ using the backward induction procedure in \Cref{eq:induction}.
For efficiency reasons, we cache the intermediate results $w(n,s)$.}
The overall backpropagation step is broken down into two phases:
\textsc{BackpropTree} iterates over every node in the tree sorted by depth in descending order\edit{, starting with the deepest nodes} (\Cref{alg:cvar-aostar-main:FFTB-sort}).
This is a significant computational effort, but it is only carried out for \edit{$s \in \mC^\Pi_{\text{new}}$ since function $w(\cdot,s)$ has not yet been evaluated for these shift values during} previous algorithm iterations.
\textsc{BackpropAncestors}, on the other hand, only \edit{evaluates $w(\cdot,s)$ with already-existing $s \in  \mC^\Pi$} at ancestors of the expanded node (akin to classical AO*).
This operation leverages \edit{cached} results from previous cycles stored in non-ancestor nodes.

\begin{figure*}[t]
    \centering
    \includegraphics[width=\linewidth]{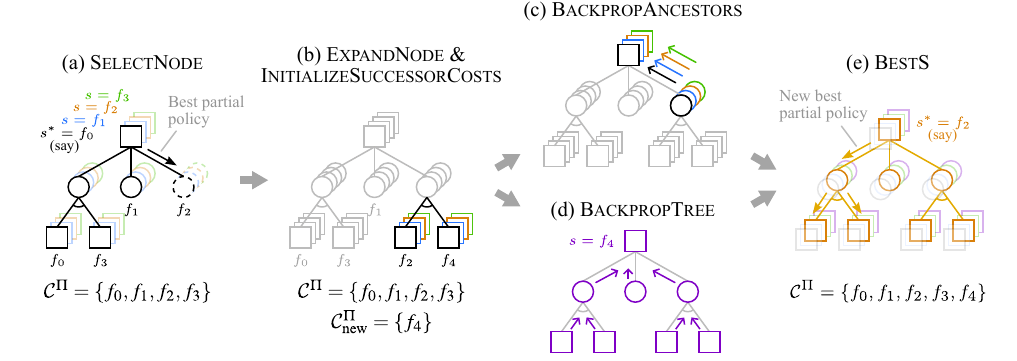}
    \caption{Conceptual illustration of one CVaR-CTP-AO* iteration.
    Each step corresponds to a routine in the main loop of~\Cref{alg:cvar-aostar-main}.
    An interpretation of this algorithm is that it grows morphologically identical AO search trees in parallel, one for each possible \edit{shift value $s \in \mC^\Pi$}.
    (a) Consider a search tree with four \edit{frontier} nodes, each assigned a unique \edit{heuristic cumulative cost $f$}.
    Let the best partial policy be the one associated with $s^* = f_0$; its \edit{(only)} non-terminal frontier node is selected for expansion.
    (b) This frontier (AND) node is expanded into two successor (OR) nodes\edit{, one of which has a new $f$-cost}, $f_4$.
    (c) A backpropagation step \edit{computes ${w(\cdot,s) \; \forall s \in \mC^\Pi}$ on the ancestors of the newly-created nodes.}
    (d) Another backpropagation \edit{through all tree nodes computes $w(\cdot,s)$} only for ${s \in \mC^\Pi_\text{new} = \{f_4\}}$.
    (e) The \edit{shift value ${s^* \in \mC^\Pi \cup \mC^\Pi_\text{new}}$ associated with the lowest CVaR is retained and the corresponding partial policy (according to $w(n_0,s^*)$) is the new best solution for the next iteration.}
    }
    \label{fig:cvar-aostar}
\end{figure*}

Lastly, \edit{$\mC^\Pi_{\text{new}}$} is merged into \edit{$\mC^\Pi$.
The best shift value $s^*$ and its corresponding partial policy (according to $w(n_0,s^*)$ where $n_0$ is the root node) associated with the lowest CVaR are found with \Cref{eq:opt_mod}; they constitute the new best partial solution.
Additional search iterations are carried out until the policy corresponding to $s^*$ has a solved root node, meaning that all of its frontier nodes correspond to terminal states.}
As illustrated in \Cref{fig:cvar-aostar}, CVaR-CTP-AO* can be interpreted as an algorithm growing a set of morphologically identical AO trees in parallel, one for each candidate $s \in \mC^\Pi$.

\subsection{Optimality of CVaR-CTP-AO*}
\label{sec:optimality}

We now provide a proof sketch establishing the optimality of CVaR-CTP-AO*.
During search, the heuristic total cost assigned to each frontier node is guaranteed not to exceed the true cost of any of its future successors, since the heuristic function $h$ is admissible.
Consequently, this implies that, for every node and every shift value $s \in \mC_\Pi$, the values computed using~\Cref{eq:inner_exp} are never overestimated.
We also know that the CVaR measure is coherent, which implies that it is monotonic~\citep[Section 3.3.1]{wang_risk-averse_2022},~\citep[Section 3]{majumdar_how_2020}.
The monotonicity property ensures that the CVaR of the partial policies found at every CVaR-CTP-AO* iteration (calculated with~\Cref{eq:opt_mod}) never overestimate the optimal CVaR.
In other words, with an admissible heuristic function, CVaR-CTP-AO* always converges to a CVaR-optimal policy from below.

If, in addition to being admissible, the heuristic function employed is consistent, \edit{the values returned by} \Cref{eq:inner_exp} for any (node, shift) \edit{input} pair will never decrease across search iterations.
By similar reasoning, the CVaR of partial policies computed with \Cref{eq:opt_mod} will monotonically converge to the optimal one over time.

\subsection{Efficiency Enhancements}

CVaR-CTP-AO* can be enhanced without compromising solution optimality.
\edit{The algorithm used in the experiments in the next section implements the efficiency mechanisms presented here.

First, our algorithm can search across an augmented state ${\widetilde{\mX} = \mX \times \mathbb{R}_{>0}}$ space that includes a positive real dimension representing the running cost incurred by the agent.
In that case, each node of the search tree represents an augmented state characterized by three components: a vertex in the graph, an information set, and the cost-to-come from the root node.
This transformation allows quick access to the cost-to-come of a frontier node $n$ without looking up the corresponding process trajectory history in the search tree (which is the behaviour of function $g(n)$ in \Cref{alg:cvar-aostar-main:init} of \Cref{alg:cvar-aostar-main}).
In fact, as shown in~\cite{bauerle_markov_2011}, this information is a sufficient statistic of the process history.
Technically, the inner optimization problem in \Cref{eq:inner_exp} over the augmented state space has an optimal substructure and admits a dynamic program.
Nonetheless, since we are using a tree-based search methodology, the cost propagation process in \Cref{alg:cvar-aostar-main} remains the same.

In this augmented version of our problem, the start state has a cost-to-come of 0 (by definition) and is defined as $\tilde{x}_0=(v_0,\mI_0,0)$.
The action space and cost function remain identical.
The augmented state transition function $\tilde{\tau}:\widetilde{\mX} \times \mA \times \widetilde{\mX} \rightarrow [0,1]$ is very similar to the original one.
Since the cost of an action $a \in \mA(\tilde{x})$ taken from some augmented state $\tilde{x}=(v,\mI,g)$ is deterministic, the cost-to-come of the resulting augmented state, $g+c(\tilde{x},a)$, is also deterministic.
For drive-observe actions, the probabilities of observation outcomes are the same as with the original state space:
\begin{multline}
    \tilde{\tau}((\text{dest}(a),\mI',g+c((v,\mI), a)) \mid (v,\mI,g), a) = \\*
    \begin{cases}
        ...& \\
        ...& \begin{aligned}
            &\quad\text{(same probabilities as} \\
            &\quad\text{in \Cref{eq:tau})},
        \end{aligned}\\
        ...&
    \end{cases}
\end{multline}
where $\tilde{x} = (v,\mI,g)$ and $\tilde{x}'=(\text{dest}(a),\mI',g+c((v,\mI), a))$ are the originating and destination augmented states, respectively.
This transformation does not change the number of CVaR-CTP-AO* iterations required to find CVaR-optimal policies; it simply extends the same search tree into a third space.
Resilience against the curse of dimensionality is another benefit of a tree-based approach over the dynamic programming algorithms iterating over the whole state space as in~\cite{bauerle_markov_2011,chow_risk-sensitive_2015}.

Second, the AO* efficiency mechanisms presented for the CTP in~\cite{aksakalli_ao_2016}, such as pruning AND nodes using worst-case cost estimates, can reduce the number of successors during the expansion of OR nodes.
Additionally, nodes of the same type representing the same augmented states (same vertex, same information set and same running cost) can be fused together to avoid expanding duplicate descendant subtrees.
In practice, searching for identical nodes has low computational overhead and is achieved through a hash table lookup, which is $O(1)$ in the best case.
Although nodes may have multiple parents with such a caching mechanism~\citep{aksakalli_ao_2016}, we will continue to refer to the data structure as an AO \textit{tree} (as opposed to an acyclic \textit{graph}) for consistency.

Third,} with minor modifications,
CVaR-CTP-AO* can also track the expected total cost \edit{(not just the values from function $w(n,s)$ for some node $n$ and shift $s$)} of partial policies.
This is achieved by also backpropagating the (untrucated) $f$-values of frontier nodes (as in AO*), which comes at the expense of greater computational overhead and memory usage in practice.
Since OR nodes represent states from which the agent makes decisions, a tie-breaking routine can be added to the \textsc{BackpropOr} procedure in favour of the child AND node with the lowest expected cost.
The overall optimization problem solved in that case is a specialized instance of the CVaR-CTP described in \Cref{eq:opt}: the policies found would have a minimal expected cost subject to the constraint that they are also CVaR-optimal, similar to the problem tackled in~\cite{rigter_planning_2022}.
This mechanism may lead to a different policy than the baseline algorithm in the presence of multiple CVaR-optimal policies.

Lastly, CVaR-CTP-AO* is an incremental algorithm in the sense that the same search instances can be leveraged for different risk aversion levels (i.e., different $\alpha$ values) to cut down on computation times for subsequent queries.
Internally, the algorithm always computes the same values with \Cref{eq:inner_exp}, resulting in the same candidate partial policies (one for each $s \in \mC^\Pi$).
The parameter $\alpha$ only dictates which of these candidate solutions should be retained as the best partial policy at each search iteration.
If a complete CVaR-optimal policy for a specific risk aversion level has not yet been found, the search simply continues where it had previously stopped.
\section{Experiments}
\label{sec:experiments}

\begin{figure*}
    \centering
    \includegraphics[width=\linewidth]{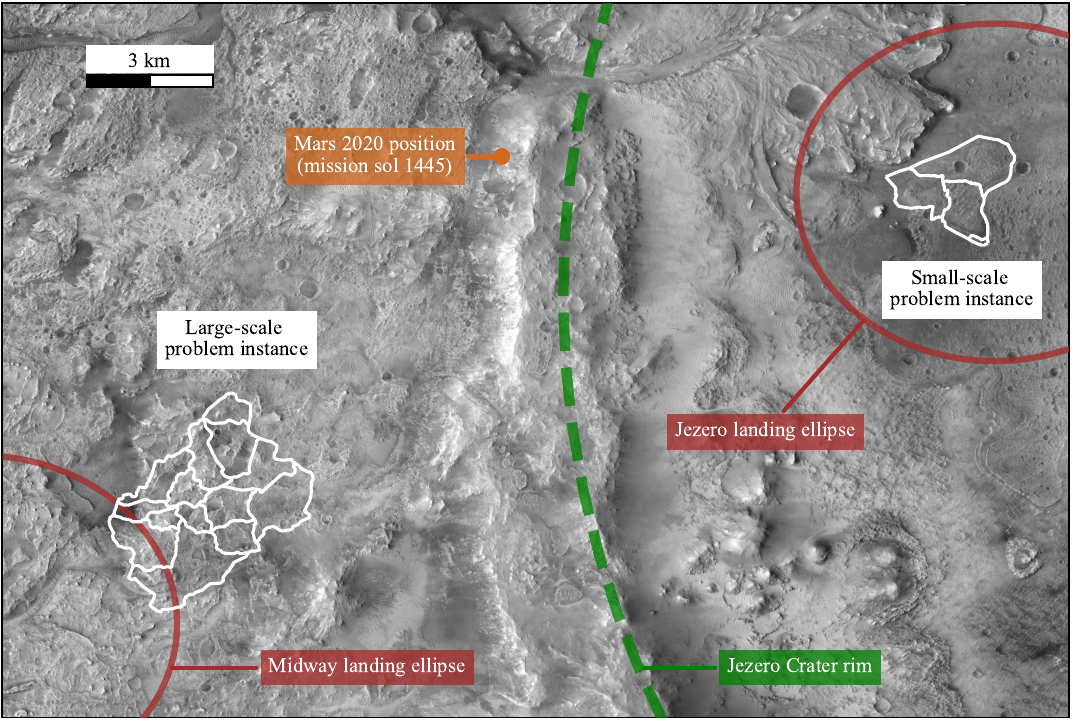}
    \caption{Global context of both simulated traverse experiments.
    The small-scale and large-scale instances use terrain information from the Jezero Crater floor and the Midway region, respectively.
    For reference, the Mars 2020 landing ellipses for both sites are shown (their positions are approximately inferred from figures in~\cite{farley_jezero-midway_2018}).
    Terrain imagery retrieved from the Mars Reconnaissance Orbiter Context Camera data archive. %
    Credit: NASA/JPL/Malin Space Science Systems.
    }
    \label{fig:global-map}
\end{figure*}

We demonstrate CVaR-CTP-AO* in two simulated global planetary mobility experiments.
Both scenarios involve graphs generated from real orbital imagery of the Martian surface.
The first problem instance involves a small graph (with a few stochastic edges) in Jezero Crater.
The second experiment involves a larger graph and simulates a traverse in the vicinity of Midway, an alternative Mars 2020 landing site next to Jezero Crater.
For both experiments, the graph generation and terrain feature extraction along every edge is accomplished with a custom QGIS tool that we open source\footnote{\finaledit{Available at \href{https://github.com/utiasSTARS/qgis-planetary-graph-tool}{https://github.com/utiasSTARS/qgis-planetary-graph-tool}}} with the current manuscript.
The global context of both experiments is shown in~\Cref{fig:global-map}.

\subsection{Small Problem Instance: Jezero Crater}

The first experiment simulates a drive near S\'{e}\'{i}tah, the region on the Jezero Crater floor with uncertain traversability previously depicted~\Cref{fig:overview}.
We illustrate in detail how CVaR-optimal policies vary as a function of risk aversion, and how information-seeking behaviours can mitigate risk when accounting for traversability correlation between edges.

\subsubsection{Problem Setup}

\begin{figure}
    \centering
    \includegraphics[width=\columnwidth]{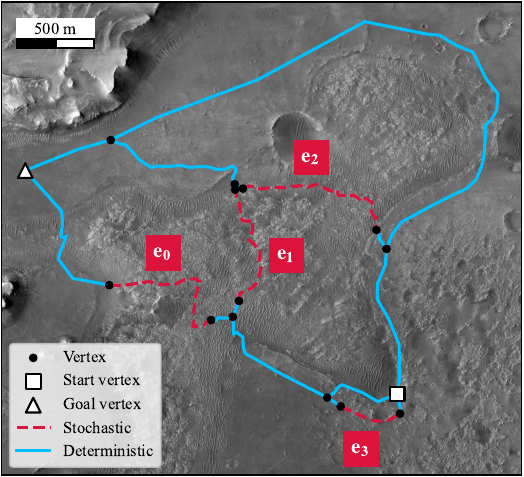}
    \caption{
    Jezero Crater floor problem instance.
    The traverse starts at the southeast of S\'{e}\'{i}tah and ends at the northwest, near the Jezero Delta.
    Terrain imagery captured with the Mars Reconnaissance Orbiter High Resolution Imaging Science Experiment (HiRISE) camera.
    Credit: NASA/JPL/University of Arizona.
    }
    \label{fig:jez-map}
\end{figure}

The graph considered in the small problem instance is shown in~\Cref{fig:jez-map}.
In this experiment, the rover starts at the southeast of the map and must traverse westward to a goal vertex near the Jezero Delta.
The graph contains four stochastic edges, three of which lie between the start and goal locations.
The fourth stochastic edge is located near the start vertex, away from the direction the rover needs to travel.
A fully deterministic (but physically longer) traverse around the uncertain terrain is also possible.

In this experiment, we minimize the total time required to reach the goal vertex.
We assume that all deterministic edges are traversable at an effective speed of 150 metres per sol\footnote{A sol is a Martian solar day (roughly 24 hours and 39 minutes, on average).}.
The effective speed along stochastic edges, however, depends on their observed traversability status (L or H).
In this demonstration, these statuses are intended to represent the level of autonomy with which a stochastic edge can be traversed.
A low-cost status (L) means that the edge is autonomously traversable at a rate of 150 metres per sol.
A high-cost status (H), on the other hand, means that the terrain is too challenging for the autonomy software.
Human operators must guide the rover through directed drives, at a significantly slower pace of 20 metres per sol.
We describe the terrain conditions along stochastic edges with a feature vector ${\boldsymbol{\phi}(e) \; \forall\, e \in E_s}$.
In the current demonstration, we consider the simple\edit{, one-dimensional case} where $\phi$ only returns a scalar representing a CFA value.
CFA models, which represent abundance of rocks of certain sizes, have been used for decades~\citep{golombek_size-frequency_1997} and have utility in rover traversability assessment~\citep{ono_mars_2018}.
We assign hypothetical CFA values to all four stochastic edges of the current problem instance: $\phi(e_{0}) = 8\%$, $\phi(e_{1}) = 10\%$, $\phi(e_{2}) = 9\%$ and $\phi(e_{3}) = 9\%$.

\begin{figure}
    \centering
    \includegraphics[width=\columnwidth]{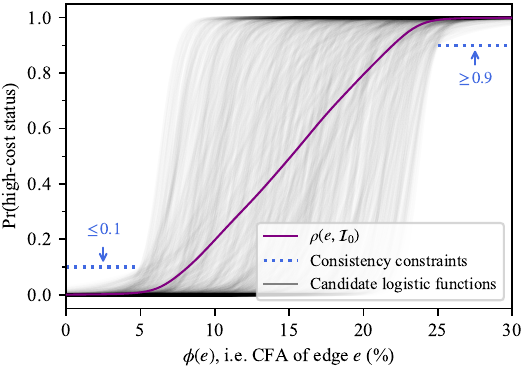}
    \caption{
    Initial probability function (purple) that an arbitrary stochastic edge $e$ has a high-cost traversability status given its assigned feature descriptor $\phi(e)$, here representing a CFA value.
    This function the weighted sum of the 1,000 candidate logistic functions shown in gray.
    All candidate functions satisfy consistency constraints enforcing low and high predicted probabilities at low (${\leq5\%}$) and high (${\geq25\%}$) CFA, respectively.
    }
    \label{fig:jez-belief}
\end{figure}

Although the CFA conveys some information about stochastic edge traversability, \edit{it remains difficult to define a specific function ${p\circ \phi: \mE_S \rightarrow [0,1]}$ assigning every edge (described by their CFA) a probability of autonomy failure as in a standard CTP instance.
This challenge is due to our limited} knowledge about rover autonomy performance in the environment explored.
\edit{It is reasonable, however, to assume that the probability of a rover autonomy failure (i.e., a high-cost status) along a stochastic edge is monotonic with respect to its CFA value.
Therefore,} we define a \textit{set} of monotonic \edit{probability} functions and assign a belief to each.
As the actual traversability statuses of stochastic edges are observed (revealed), this belief is gradually refined using Bayes' rule\edit{, akin to a Bayes-Adaptive MDP}.
Specifically, in this experiment we sample a set of 1,000 candidate logistic functions\footnote{Each \edit{probability function is a logistic} of the form $p(x)=1/(1+e^{-a(x-b)})$ with different $a$ and $b$ hyperparameters. We choose this formula because the output is bounded between 0 and 1 (ideal to represent probabilities) and monotonic with respect to the input (in this case, a CFA value).} (which we represent with \edit{$P$}) that are consistent with \edit{our} basic assumptions about probable outcomes at low and high CFA.
These consistency constraints and the set of sampled logistic functions are visualized in \Cref{fig:jez-belief}.

Initially, the actual traversability status of all four stochastic edges is unknown.
The information component of the rover's start state is $\mI_0=\{\text{A},\text{A},\text{A},\text{A}\}$ and all 1,000 candidate logistic functions \edit{${p \in P}$} are assigned an equal prior belief ${b_{\mI_0}(p) = 1e^{-3}}$.
Given an arbitrary information set $\mI$, the posterior belief over candidate logistic functions ${p \in P}$ is computed with the Bayesian formula
\begin{equation} \label{eq:belief-update}
    b_{\mI}(p) = b(p \mid \mI) \propto \text{Bern}(\mI \mid p)^\theta b_{\mI_0}(p),
\end{equation}
where $\text{Bern}(\mI \mid p)^\theta$ is the weighted Bernoulli likelihood of observations in $\mI$ with function $p$.
All the results that follow are obtained with $\theta=5$.
Finally, we define the probability that an unobserved stochastic edge $e$ has a high-cost status (H) as the sum of all logistic function candidates weighted by their belief:
\begin{equation} \label{eq:rho}
    \rho(e,\mI)\vcentcolon=\sum_p^P b_{\mI}(p)p(\phi(e)).
\end{equation}
The initial probability function, $\rho(\cdot,\mI_0)$, is also plotted in \Cref{fig:jez-belief}.

\subsubsection{Results}

\begin{figure}
    \centering
    \includegraphics[width=\columnwidth]{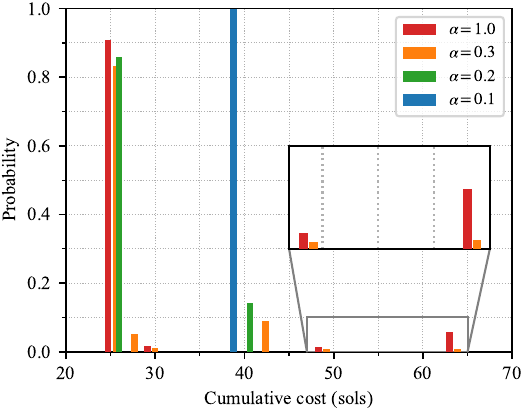}
    \caption{
    Cost distribution of four CVaR-optimal policies for the small-scale experiment.
    The risk-agnostic policy (${\alpha=1.0}$) has costly worst-case outcomes with non-negligible probabilities.
    The policy corresponding to $\alpha=0.3$ has slightly worst worst-case outcomes, but with a much lower probabilities.
    At $\alpha=0.1$, the optimal policy is the path that avoids all stochastic edges, leading to a deterministic outcome.
    }
    \label{fig:small-costs}
\end{figure}

We investigate four optimal policies obtained with the CVaR-CTP-AO* algorithm at different risk aversion levels: $\alpha = 1.0$ (i.e., the risk-agnostic case), $0.3$, $0.2$, and $0.1$.
The cumulative cost distribution of each policy is shown in~\Cref{fig:small-costs}.
As a sanity check,~\Cref{tab:jez-cvar} shows the CVaR$_{\alpha=1.0}$, CVaR$_{0.3}$, CVaR$_{0.2}$ and CVaR$_{0.1}$ of all policies and demonstrates that each strategy has the lowest \edit{CVaR} value for the $\alpha$ it optimizes for.

\begin{figure*}
    \centering
    \includegraphics[width=\linewidth]{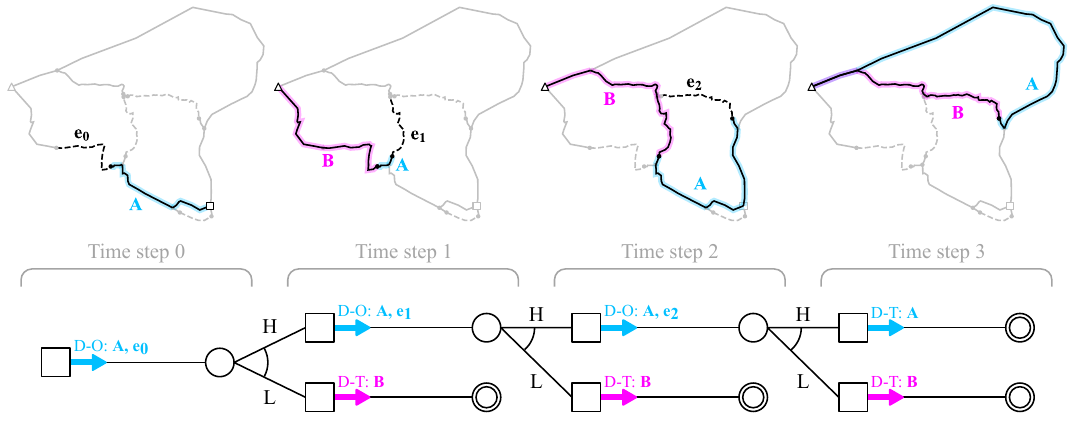}
    \caption{
    Optimal risk-agnostic policy tree ($\alpha=1.0$) for the small-scale experiment, stretching from left to right.
    At each time step, all possible OR nodes (rectangular nodes) are labelled with optimal actions, which are illustrated in the corresponding graph diagram above.
    D-O (drive-observe) actions indicate the deterministic path driven and stochastic edge observed, and D-T (drive-terminate) actions indicate the deterministic path followed to the goal vertex.}
    \label{fig:T_a10}
\end{figure*}

The optimal risk-agnostic policy ($\alpha=1.0$) is illustrated in~\Cref{fig:T_a10}.
This policy is optimistic and begins with a drive straight to the stochastic edge $e_0$ and then to edge $e_1$ if the former has a high-cost (H) traversability status.
If both $e_0$ and $e_1$ have a high-cost status, the rover returns to the start vertex and continues to $e_2$.
In the worst-case (if $e_2$ also has a high-cost status), the agent follows the long determistic path to the goal vertex.

\begin{table}
    \centering
    \caption{CVaR at different risk aversion levels of all policies for the small-scale experiment. Each policy minimizes the level it optimizes for.}
    \begin{tabular}{p{1.2cm}llll}
    \toprule
    Policy & $\cvar_{1.0}$ &$\cvar_{0.3}$ &$\cvar_{0.2}$ &$\cvar_{0.1}$ \\
    \midrule
    $\pi_{1.0}$ &   \textbf{27.38}  & 33.60             & 38.04             & 51.37\\
    $\pi_{0.3}$ &   27.74           & \textbf{32.74}    & 36.30             & 44.61\\
    $\pi_{0.2}$ &   28.03           & 32.84             & \textbf{36.28}    & 40.64\\
    $\pi_{0.1}$ &   38.75           & 38.75             & 38.75             & \textbf{38.75}\\
    \bottomrule
    \end{tabular}
    \label{tab:jez-cvar}
\end{table}

\begin{figure*}
    \centering
    \includegraphics[width=\linewidth]{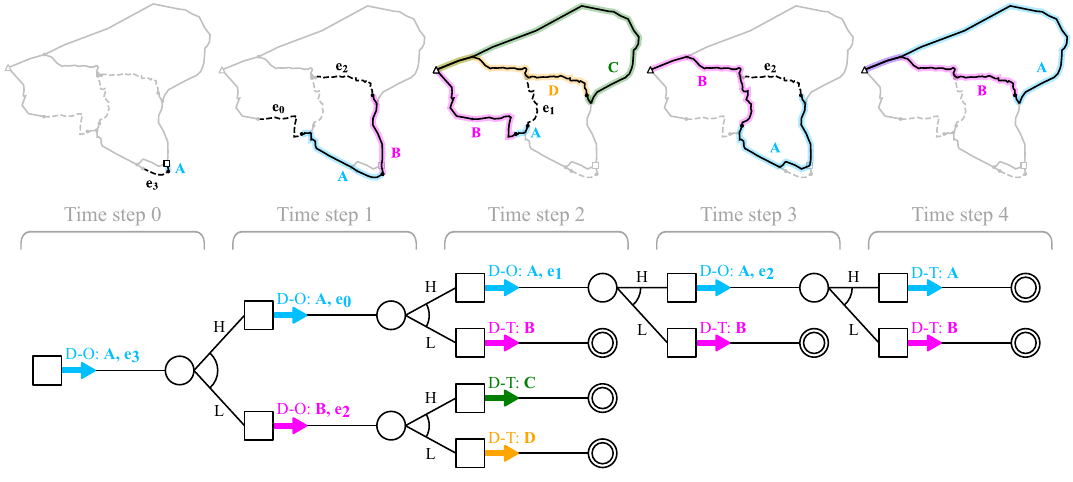}
    \caption{
    Optimal policy tree for the small-scale experiment obtained with $\alpha=0.3$, stretching from left to right.
    At each time step, all possible OR nodes (rectangular nodes) are labelled with optimal actions, which are illustrated in the corresponding graph diagram above.
    D-O (drive-observe) actions indicate the deterministic path driven and edge observed, and D-T (drive-terminate) actions indicate the path followed to reach the goal vertex.
    }
    \label{fig:T_a03}
\end{figure*}

With $\alpha=0.3$, the policy found is more risk-averse: it first takes the rover on a very small detour to observe edge $e_3$, as illustrated in \Cref{fig:T_a03}.
If this edge has a low-cost status, then the rover commits to $e_0$ and follows the same decision-making scheme as the risk-agnostic policy from that point onward.
Otherwise, the rover heads straight to edge $e_2$ and, if this edge has a high-cost status, the agent follows the long route to the goal.
It is important to note that with such a policy, the worst-case outcome \edit{has a greater cost} than with the risk-agnostic policy: it corresponds to the (unlikely) event that edge $e_3$ has a low-cost status and all other stochastic edges have a high-cost status.
However, as illustrated in \Cref{fig:small-costs}, this \edit{undesirable} outcome is significantly less probable than the worst-case outcome of the risk-agnostic policy.
The policy computed with $\alpha=0.3$ has a lower (and optimal) CVaR$_{0.3}$ value.
We also highlight that our algorithm and problem setup do not have an explicit information-seeking objective.
The detour to $e_3$ emerges as a risk-mitigation behaviour \edit{that} is implicitly enabled by accounting for traversability correlations in the environment.

The CVaR$_{0.2}$-optimal policy guides the rover straight to edge $e_2$ and along the long detour in the worst case.
Lastly, setting ${\alpha=0.1}$ in this experiment is akin to using a worst-case policy (which assumes that every stochastic edge has a high-cost status) and guides the rover along the long deterministic path from the \edit{very} start.

\subsection{Large Problem Instance: Midway}

\begin{figure}
    \centering
    \includegraphics[width=\columnwidth]{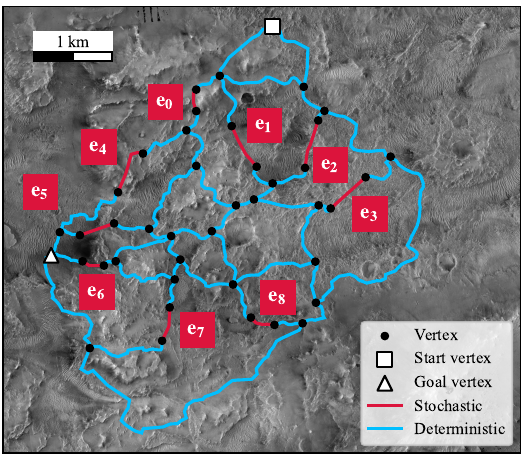}
    \caption{
    Midway region problem instance.
    Terrain imagery captured with the Mars Reconnaissance Orbiter High Resolution Imaging Science Experiment (HiRISE) camera.
    Credit: NASA/JPL/University of Arizona.
    }
    \label{fig:mid-map}
\end{figure}

The second experiment is inspired by a potential Mars 2020 mission extension scenario~\citep{farley_jezero-midway_2018} and simulates a rover approaching the Midway region.
The graph of this new problem instance, shown in~\Cref{fig:mid-map}, is larger than the previous one.

\subsubsection{Problem Setup}

\begin{figure}
    \centering
    \includegraphics[width=\columnwidth]{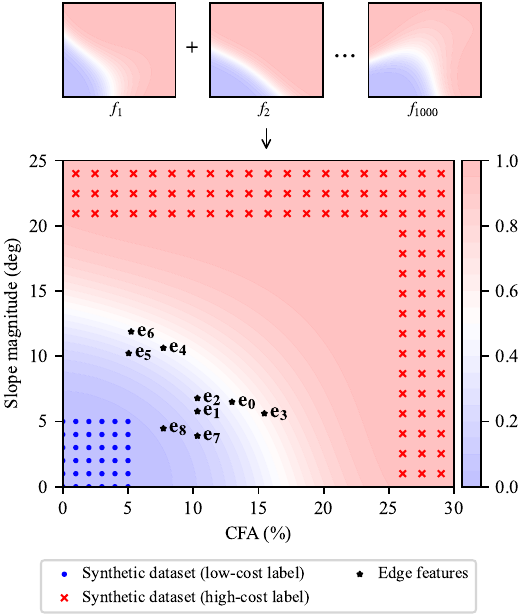}
    \caption{
    Initial function $\rho(\cdot,\mI_0)$ expressing the probability of a high-cost status over the two-dimensional space of edge features for the large-scale experiment.
    This surface is the weighted sum of 1,000 candidate functions sampled from a two-dimensional Gaussian process classifier, three of which are depicted at the top.
    The classifier's parameters, listed in~\Cref{tab:gpc}, are optimized with the synthetic binary dataset shown.
    The features of all nine stochastic edges in the graph of the current problem instance are also visualized.
    }
    \label{fig:mdw-belief}
\end{figure}

The graph contains nine stochastic edges in the general area between the start and goal vertices.
A fully deterministic (but very long) path to the goal vertex is also available.
In this problem instance, the feature descriptor ${\boldsymbol{\phi}(e) \; \forall e \in E_s}$ is two-dimensional: it associates every stochastic edge with a CFA value and a terrain slope magnitude indicator.
These features are calculated as follows.
Slope and CFA profiles along every stochastic edge $e$ are generated by sampling, at 3 metres intervals, the Midway slope and CFA maps of a preliminary Mars 2020 global traversability analysis~\citep{ono_mars_2018}.
Then, one of the extracted slope-CFA pairs is retained as the representative feature set $\boldsymbol{\phi}(e)$.
This approach provides more representative edge descriptors than computing a summary statistic of each feature profile independently.
In this experiment, we retain the slope-CFA pair that is the closest to the median of the sum of both features.\footnote{
Our selection criterion combines both features to jointly account for them. We use a sum since slope and CFA values have a similar magnitude and non-negative range, and choose the median for its resilience against outliers in profile data.}
Each stochastic edge feature is plotted in~\Cref{fig:mdw-belief}.

Once again, the cost to minimize is the total traverse time to the goal vertex.
The effective traverse speed along deterministic edges and stochastic edges with a low-cost status (L) is assumed to be 150 metres per sol.
Unlike the previous demonstration, now stochastic edges with a high-cost status (H) have an infinite cost.
In this experiment, a high-cost status observation represents treacherous terrain that should be entirely avoided for rover safety.

Function $\rho$ \edit{works the same way} as the one in \Cref{eq:rho,eq:belief-update} for the first experiment\edit{, except that here its input is two-dimensional.}
We again assume that it relies on a set $P$ of 1,000 candidate functions mapping the (two-dimensional) edge feature space to a probability of a high-cost status.
Deciding on the parametrization of these candidate functions is however nontrivial.
In the previous (small-scale) problem instance, the logistic function was a sensible choice, given the \edit{monotonic} relationship between CFA and ease of terrain traversability.
In the current problem instance, however, it is unclear how CFA and slope jointly affect traversability and what formula should represent this relationship.
This representation ambiguity worsens in higher dimensions; it is an open problem in adaptive mobility for unknown field environments and finding the best approach falls outside the scope of the current paper.
In this experiment, we nevertheless propose an implementation to overcome this challenge.
We circumvent the need to choose a specific formula for candidate functions $P$ and instead adopt a \textit{non-parametric} method to sample them.

Using the GPy library~\citep{gpy_gpy_2012}, we tune a Gaussian process classifier (GPC) using a synthetic binary classification dataset that represent our traversability assumptions at both ends of the feature space: we assume that flat terrain with low CFA is generally traversable (low-cost status) while steep or high CFA terrain is untraversable (high-cost status).
This dataset is visualized in~\Cref{fig:mdw-belief} and the optimized GPC configuration is detailed in~\Cref{tab:gpc}.

\begin{table}
    \centering
    \caption{Gaussian process binary classifier configuration with the GPy library~\citep{gpy_gpy_2012}.}
    \begin{tabular}{ll}
    \toprule
    \textbf{Component} &
    \textbf{Configuration/Value}\\
    \midrule
    Mean function                   & None (zero function) \\
    Kernel function type            & Squared-exponential \\
    \begin{tabular}{@{}l@{}}Tuned kernel lengthscales \\ (CFA \& slope dimensions)\end{tabular}        & (1.68, 1.62) \\
    Tuned kernel variance           & 6.38 \\
    Inference approximation         & Expectation-propagation \\
    Likelihood, link function       & Bernoulli, probit \\
    \bottomrule
    \end{tabular}
    \label{tab:gpc}
\end{table}

Each candidate function is generated by sampling the GPC \edit{on} a grid of discrete points spanning the two-dimensional feature space\footnote{The GPC is sampled in latent (Gaussian) space and the outputs are mapped to probability space using the classifier's likelihood link function (the probit function, in this case).} and converting these discrete values into a continuous probability surface using bilinear interpolation.
For consistency, a function is added to the set $P$ only if the error between its predicted probabilities and the synthetic dataset labels is less than 0.1.
\Cref{fig:mdw-belief} shows the synthetic dataset, a few functions from the set $P$ and the initial probability surface $\rho(\cdot, \mI_0)$ obtained with \Cref{eq:rho} across the two-dimensional edge feature space.

\subsubsection{Results}

We again use CVaR-CTP-AO* to find optimal policies at different risk-aversion levels.
We pay a special attention to those found with $\alpha=1.0$ (risk-agnostic case), $0.5$, $0.4$ and $0.3$ as they correspond to distinct decision-making schemes.
Their cost distributions are shown in~\Cref{fig:large-costs}.

A rover following the risk-agnostic policy first tries to traverse stochastic edges $e_0$ and $e_4$.
Then, if unsuccessful, it observes other edges while gradually moving towards the southeast portion of the graph.
If none of the observed edges are traversable (i.e., have a high-cost status), the rover falls back to the long deterministic traverse.
The optimal policy with $\alpha=0.5$ skips edge $e_4$ to avoid detours early during the traverse, and the optimal policy with $\alpha=0.4$ avoids both edges $e_0$ and $e_4$ altogether.
These increasingly risk-averse behaviours, however, come at the cost of worsening best-case outcomes as shown in~\Cref{fig:large-costs}.

\begin{figure}
    \centering
    \includegraphics[width=\columnwidth]{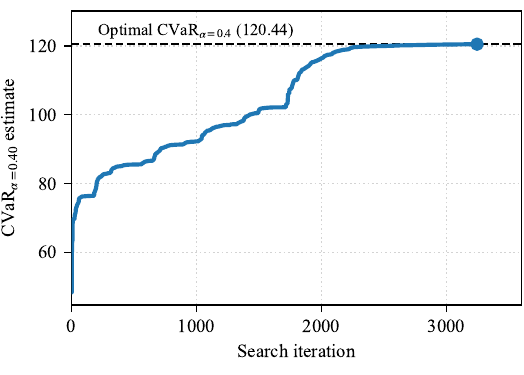}
    \caption{
    Evolution of the best $\text{CVaR}_{\alpha=0.4}$ estimate during search.
    Our consistent heuristic function ensures that the estimate converges monotonically to the optimal value without overestimating it.
    }
    \label{fig:mdw-cvar}
\end{figure}

\begin{figure*}
    \centering
    \includegraphics[width=\textwidth]{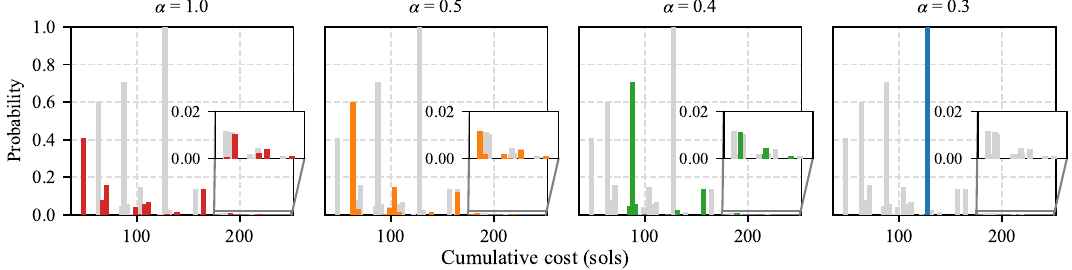}
    \caption{
    Cost distribution of four CVaR-optimal policies (with $\alpha = 1.0$, $0.5$, $0.4$ and $0.3$) for the large problem instance in the vicinity of Midway.
    For ease of comparison, the shadow of all other distributions is shown in the background in each plot.
    The zoomed-in windows show low-probability ($<2\%$) outcomes with costs between 175 and 250 sols.
    }
    \label{fig:large-costs}
\end{figure*}

\Cref{fig:mdw-cvar} shows the evolution of the CVaR of the best partial policy found at every search cycle (the objective function of~\Cref{eq:opt_mod} evaluated at $s=s^*$) with $\alpha=0.4$.
As described in~\Cref{sec:optimality}, the use of an admissible heuristic function with CVaR-CTP-AO* ensures that this quantity never overestimates the optimal value.
Our heuristic is also consistent, which explains the monotonic convergence to the optimal solution.

\finaledit{
\subsubsection{Value of Information}

To highlight the benefit of accounting for edge traversability correlations in the previous Midway experiment, we run CVaR-CTP-AO* on a slightly different problem instance.
We keep the initial traversability probability of every edge fixed for the duration of the traversal, consistent with the standard Canadian Traveller Problem formulation.
In this ``uncorrelated'' case, the $\text{CVaR}_\alpha$-optimal solutions for $\alpha = 1.0$, $0.5$ and $0.4$ happen to be the same policy tree with 26 leaf (terminal) nodes.
The corresponding cost distribution has an entropy of 2.87 bits.
In contrast, in the previous ``correlated'' case, the policies found for the same risk aversion levels contain at most 16 leaf nodes, meaning that the corresponding cost distributions (shown in \Cref{fig:large-costs}) are consistently less spread out.
The entropy values of these distributions are 2.66 bits, 1.95 bits, and 1.52 bits for $\alpha = 1.0$, $0.5$, and $0.4$, respectively.
With $\alpha = 0.3$, the same deterministic worst-case policy is obtained for both the uncorrelated and correlated problem instances (with an entropy of 0 bits).
}
\section{Conclusion}

In this paper, we defined a novel risk-averse optimization problem, CVaR-CTP, tailored to the adaptive traversal of uncertain planetary environments modelled as graphs with stochastic edge costs.
This particular task seeks policies minimizing a conditional value-at-risk criterion (CVaR), which is a risk measure with an intuitive interpretation.
We formulated the CVaR-CTP-AO* algorithm to find exact and optimal policies using a forward search paradigm.
Lastly, we validated our approach through simulated traverses of graphs generated from real orbital maps of the Martian surface.
Our experimental setup took advantage of terrain properties to correlate traversability costs between different stochastic edges; we empirically showed that in such settings, information-seeking behaviours are a form of risk-aversion.

Our approach could be improved in many ways in follow-up work.
Since the size of the search space increases exponentially with the number of stochastic edges in the graph, CVaR-CTP-AO* cannot scale to very large graphs, as it only returns complete solutions.
An `online' variant of the algorithm, which limits the search tree depth and accepts multiple queries during a single traverse trial, as presented by~\cite{guo_robust_2019}, would overcome this limitation at the expense of solution optimality.
Furthermore, accounting for the full context of terrain regions mapped to graph edges (as opposed to summarized quantities as in the current work) would allow much more complex adaptive decision-making schemes.
In reality, traversing a single edge in a global graph results in the collection of a large, diverse set of observations of the environment at a local level.
A more sophisticated traversability belief update mechanism would be needed to account for all of these data points and to provide a more nuanced traversability assessments, that is, beyond the binary statuses considered in the current work.

\edit{
An essential aspect of strategic mobility planning in planetary environments is identifying dangerous terrain that should be avoided entirely before constructing a network of candidate global routes (represented as a planning graph in this work).
These high-level heuristics, typically derived from terrain features, and have significant implications for rover safety.
Steep slopes, terrain with high CFA, and fields of large ripples are examples of mobility hazards that current Mars rover missions avoid~\citep{arvidson_mars_2017,ono_mars_2018}.
Local mobility observations can influence how strictly these high-level heuristics are applied and may warrant modifications to the planning graph (updated layout with new edges and vertices).
Consequently, an interesting research direction in risk-averse strategic planetary mobility is to account for graph topology updates in addition to evolving traversability beliefs along existing routes (graph edges).
}

\section*{Acknowledgements}

We are deeply grateful to Dr. Marc Rigter and Professor Margaret Chapman for valuable insights about the CVaR risk measure in optimization frameworks.

\section*{Declarations}

This work was supported by the Natural Sciences and Engineering Research Council of Canada through the Discovery Grant program and the Canada Graduate Scholarships (Doctoral) program.
The authors declare that they have no conflict of interest.

\bibliography{references.bib}

@article{aksakalli_ao_2016,
  author = {Aksakalli, Vural and Sahin, O. Furkan and Ari, Ibrahim},
  doi = {10.1287/ijoc.2015.0668},
  issn = {1091-9856},
  journal = {INFORMS Journal on Computing},
  month = {February},
  number = {1},
  pages = {96--111},
  title = {An {AO}* {Based} {Exact} {Algorithm} for the {Canadian} {Traveler} {Problem}},
  urldate = {2022-04-28},
  volume = {28},
  year = {2016}
}

@article{arvidson_mars_2017,
  author = {Arvidson, Raymond E. and Iagnemma, Karl D. and Maimone, Mark and Fraeman, Abigail A. and Zhou, Feng and Heverly, Matthew C. and Bellutta, Paolo and Rubin, David and Stein, Nathan T. and Grotzinger, John P. and Vasavada, Ashwin R.},
  copyright = {© 2016 Wiley Periodicals, Inc.},
  doi = {https://doi.org/10.1002/rob.21647},
  issn = {1556-4967},
  journal = {Journal of Field Robotics},
  language = {en},
  number = {3},
  pages = {495--518},
  title = {Mars {Science} {Laboratory} {Curiosity} {Rover} {Megaripple} {Crossings} up to {Sol} 710 in {Gale} {Crater}},
  urldate = {2021-01-25},
  volume = {34},
  year = {2017}
}

@article{arvidson_relating_2017,
  author = {Arvidson, R. E. and DeGrosse, P. and Grotzinger, J. P. and Heverly, M. C. and Shechet, J. and Moreland, S. J. and Newby, M. A. and Stein, N. and Steffy, A. C. and Zhou, F. and Zastrow, A. M. and Vasavada, A. R. and Fraeman, A. A. and Stilly, E. K.},
  doi = {10.1016/j.jterra.2017.03.001},
  issn = {0022-4898},
  journal = {Journal of Terramechanics},
  language = {en},
  month = {October},
  pages = {73--93},
  series = {Manned/{Unmanned} {Ground} {Vehicles}: {Off}-{Road} {Dynamics} and {Mobility}},
  title = {Relating geologic units and mobility system kinematics contributing to {Curiosity} wheel damage at {Gale} {Crater}, {Mars}},
  urldate = {2021-01-22},
  volume = {73},
  year = {2017}
}

@article{bauerle_markov_2011,
  author = {Bäuerle, Nicole and Ott, Jonathan},
  doi = {10.1007/s00186-011-0367-0},
  issn = {1432-5217},
  journal = {Mathematical Methods of Operations Research},
  language = {en},
  month = {December},
  number = {3},
  pages = {361--379},
  title = {Markov {Decision} {Processes} with {Average}-{Value}-at-{Risk} criteria},
  urldate = {2024-02-13},
  volume = {74},
  year = {2011}
}

@inproceedings{bnaya_canadian_2009,
  author = {Bnaya, Zahy and Felner, Ariel and Shimony, Solomon Eyal},
  booktitle = {Twenty-{First} {International} {Joint} {Conference} on {Artificial} {Intelligence} ({IJCAI})},
  pages = {437--442},
  title = {Canadian {Traveler} {Problem} with {Remote} {Sensing}.},
  urldate = {2024-04-29},
  year = {2009}
}

@article{cai_evora_2024,
  author = {Cai, Xiaoyi and Ancha, Siddharth and Sharma, Lakshay and Osteen, Philip R. and Bucher, Bernadette and Phillips, Stephen and Wang, Jiuguang and Everett, Michael and Roy, Nicholas and How, Jonathan P.},
  doi = {10.1109/TRO.2024.3431828},
  issn = {1941-0468},
  journal = {IEEE Transactions on Robotics},
  pages = {3756--3777},
  shorttitle = {{EVORA}},
  title = {{EVORA}: {Deep} {Evidential} {Traversability} {Learning} for {Risk}-{Aware} {Off}-{Road} {Autonomy}},
  urldate = {2024-09-09},
  volume = {40},
  year = {2024}
}

@inproceedings{cao_anmip_2023,
  author = {Cao, Chenyang and Xu, Xujun and Gong, Xiaofei and Lu, Bo and Chi, Wenzheng and Sun, Lining},
  booktitle = {2023 {IEEE} {International} {Conference} on {Robotics} and {Biomimetics} ({ROBIO})},
  doi = {10.1109/ROBIO58561.2023.10354912},
  month = {December},
  pages = {1--7},
  shorttitle = {{ANMIP}},
  title = {{ANMIP}: {Adaptive} {Navigation} based on {Mutual} {Information} {Perception} in {Uncertain} {Environments}},
  urldate = {2025-04-06},
  year = {2023}
}

@inproceedings{chow_algorithms_2014,
  author = {Chow, Yinlam and Ghavamzadeh, Mohammad},
  booktitle = {Advances in {Neural} {Information} {Processing} {Systems}},
  publisher = {Curran Associates, Inc.},
  title = {Algorithms for {CVaR} {Optimization} in {MDPs}},
  urldate = {2022-11-10},
  volume = {27},
  year = {2014}
}

@inproceedings{chow_risk-sensitive_2015,
  author = {Chow, Yinlam and Tamar, Aviv and Mannor, Shie and Pavone, Marco},
  booktitle = {Advances in {Neural} {Information} {Processing} {Systems}},
  publisher = {Curran Associates, Inc.},
  shorttitle = {Risk-{Sensitive} and {Robust} {Decision}-{Making}},
  title = {Risk-{Sensitive} and {Robust} {Decision}-{Making}: a {CVaR} {Optimization} {Approach}},
  urldate = {2022-11-10},
  volume = {28},
  year = {2015}
}

@article{dixit_step_2024,
  author = {Dixit, Anushri and Fan, David D. and Otsu, Kyohei and Dey, Sharmita and Agha-Mohammadi, Ali-Akbar and Burdick, Joel},
  doi = {10.55417/fr.2024006},
  journal = {Field Robotics},
  pages = {182--210},
  title = {{STEP}: {Stochastic} {Traversability} {Evaluation} and {Planning} for {Risk}-{Aware} {Navigation}; {Results} {From} the {DARPA} {Subterranean} {Challenge}},
  volume = {4},
  year = {2024}
}

@inproceedings{duff_design_2003,
  author = {Duff, Michael O.},
  booktitle = {Proceedings of the 20th {International} {Conference} on {Machine} {Learning} ({ICML}-03)},
  pages = {131--138},
  title = {Design for an optimal probe},
  urldate = {2025-07-23},
  year = {2003}
}

@inproceedings{endo_risk-aware_2023,
  author = {Endo, Masafumi and Taniai, Tatsunori and Yonetani, Ryo and Ishigami, Genya},
  booktitle = {2023 {IEEE} {International} {Conference} on {Robotics} and {Automation} ({ICRA})},
  doi = {10.1109/ICRA48891.2023.10161466},
  month = {May},
  pages = {11852--11858},
  title = {Risk-aware {Path} {Planning} via {Probabilistic} {Fusion} of {Traversability} {Prediction} for {Planetary} {Rovers} on {Heterogeneous} {Terrains}},
  urldate = {2024-07-12},
  year = {2023}
}

@inproceedings{farley_jezero-midway_2018,
  address = {Gendale, California},
  author = {Farley, Ken and Stack Morgan, Katie and Williford, Ken},
  booktitle = {Fourth {Landing} {Site} {Selection} {Workshop} for {Mars} 2020},
  month = {October},
  title = {Jezero-{Midway} {Interellipse} {Traverse} {Mission} {Concept}},
  year = {2018}
}

@inproceedings{ferguson_pao_2004,
  author = {Ferguson, D. and Stentz, A. and Thrun, S.},
  booktitle = {{IEEE} {International} {Conference} on {Robotics} and {Automation}, 2004. {Proceedings}. {ICRA} '04. 2004},
  doi = {10.1109/ROBOT.2004.1307491},
  month = {April},
  pages = {2840--2847 Vol.3},
  title = {{PAO} for planning with hidden state},
  urldate = {2024-02-07},
  volume = {3},
  year = {2004}
}

@article{golombek_size-frequency_1997,
  author = {Golombek, M. and Rapp, D.},
  doi = {10.1029/96JE03319},
  issn = {2156-2202},
  journal = {Journal of Geophysical Research: Planets},
  language = {en},
  number = {E2},
  pages = {4117--4129},
  shorttitle = {Size-frequency distributions of rocks on {Mars} and {Earth} analog sites},
  title = {Size-frequency distributions of rocks on {Mars} and {Earth} analog sites: {Implications} for future landed missions},
  urldate = {2021-10-08},
  volume = {102},
  year = {1997}
}

@misc{gpy_gpy_2012,
  author = {{GPy}},
  title = {{GPy}: {A} {Gaussian} process framework in python},
  url = {http://github.com/SheffieldML/GPy},
  year = {2012}
}

@article{guo_dual_2022,
  author = {Guo, Hongliang and Shi, Rui and Rus, Daniela and Yau, Wei-Yun},
  doi = {10.1109/TVT.2022.3191490},
  issn = {1939-9359},
  journal = {IEEE Transactions on Vehicular Technology},
  month = {November},
  number = {11},
  pages = {11465--11479},
  title = {Dual {Dynamic} {Programming} for the {Mean} {Standard} {Deviation} {Canadian} {Traveller} {Problem}},
  urldate = {2024-04-29},
  volume = {71},
  year = {2022}
}

@inproceedings{guo_robust_2019,
  author = {Guo, H. and Barfoot, T. D.},
  booktitle = {2019 {International} {Conference} on {Robotics} and {Automation} ({ICRA})},
  doi = {10.1109/ICRA.2019.8794252},
  month = {May},
  pages = {5523--5529},
  title = {The {Robust} {Canadian} {Traveler} {Problem} {Applied} to {Robot} {Routing}},
  year = {2019}
}

@article{howard_risk-sensitive_1972,
  author = {Howard, Ronald A. and Matheson, James E.},
  journal = {Management science},
  number = {7},
  pages = {356--369},
  title = {Risk-sensitive {Markov} decision processes},
  volume = {18},
  year = {1972}
}

@inproceedings{huang_stochastic_2023,
  author = {Huang, Yizhou and Dugmag, Hamza and Barfoot, Timothy D. and Shkurti, Florian},
  booktitle = {2023 {IEEE} {International} {Conference} on {Robotics} and {Automation} ({ICRA})},
  doi = {10.1109/ICRA48891.2023.10160894},
  month = {May},
  pages = {1055--1061},
  title = {Stochastic {Planning} for {ASV} {Navigation} {Using} {Satellite} {Images}},
  urldate = {2024-04-29},
  year = {2023}
}

@article{koenig_fast_2005,
  author = {Koenig, S. and Likhachev, M.},
  doi = {10.1109/TRO.2004.838026},
  issn = {1941-0468},
  journal = {IEEE Transactions on Robotics},
  month = {June},
  number = {3},
  pages = {354--363},
  title = {Fast replanning for navigation in unknown terrain},
  volume = {21},
  year = {2005}
}

@inproceedings{lamarre_importance_2024,
  address = {Brisbane, Australia},
  author = {Lamarre, Olivier and Kelly, Jonathan},
  booktitle = {2024 {International} {Symposium} on {Artificial} {Intelligence}, {Robotics} and {Automation} in {Space}},
  copyright = {All rights reserved},
  doi = {10.48550/arXiv.2409.19455},
  month = {November},
  title = {The {Importance} of {Adaptive} {Decision}-{Making} for {Autonomous} {Long}-{Range} {Planetary} {Surface} {Mobility}},
  urldate = {2024-12-19},
  year = {2024}
}

@inproceedings{lim_shortest_2017,
  author = {Lim, Zhan Wei and Hsu, David and Lee, Wee Sun and Sun, W.},
  booktitle = {{UAI}},
  shorttitle = {Shortest {Path} under {Uncertainty}},
  title = {Shortest {Path} under {Uncertainty}: {Exploration} versus {Exploitation}.},
  year = {2017}
}

@incollection{majumdar_how_2020,
  address = {Cham},
  author = {Majumdar, Anirudha and {Marco Pavone}},
  booktitle = {Robotics {Research}: {The} 18th {International} {Symposium} {ISRR}},
  doi = {10.1007/978-3-030-28619-4_10},
  isbn = {978-3-030-28618-7 978-3-030-28619-4},
  language = {en},
  pages = {75--84},
  publisher = {Springer International Publishing},
  shorttitle = {How {Should} a {Robot} {Assess} {Risk}?},
  title = {How {Should} a {Robot} {Assess} {Risk}? {Towards} an {Axiomatic} {Theory} of {Risk} in {Robotics}},
  urldate = {2020-08-29},
  volume = {10},
  year = {2020}
}

@inproceedings{martelli_additive_1973,
  address = {San Francisco, CA, USA},
  author = {Martelli, A. and Montanari, U.},
  booktitle = {Proceedings of the 3rd international joint conference on {Artificial} intelligence},
  month = {August},
  pages = {1--11},
  publisher = {Morgan Kaufmann Publishers Inc.},
  series = {{IJCAI}'73},
  title = {Additive {AND}/{OR} graphs},
  urldate = {2024-01-22},
  year = {1973}
}

@book{nilsson_principles_1982,
  author = {Nilsson, Nils J},
  publisher = {Springer Science \& Business Media},
  title = {Principles of {Artificial} {Intelligence}},
  year = {1982}
}

@misc{noauthor_nasas_2024,
  author = {NASA},
  journal = {NASA Jet Propulsion Laboratory (JPL)},
  language = {en-US},
  month = {June},
  title = {{NASA}'s {Perseverance} {Fords} {An} {Ancient} {River} {To} {Reach} {Science} {Target}},
  url = {https://www.jpl.nasa.gov/news/nasas-perseverance-fords-an-ancient-river-to-reach-science-target/},
  urldate = {2025-06-20},
  year = {2024}
}

@inproceedings{ono_mars_2018,
  author = {Ono, Masahiro and Heverly, Matthew and Rothrock, Brandon and Almeida, Eduardo and Calef, Fred and Soliman, Tariq and Williams, Nathan and Gengl, Hallie and Ishimatsu, Takuto and Nicholas, Austin},
  booktitle = {2018 {AIAA} {SPACE} and {Astronautics} {Forum} and {Exposition}},
  pages = {5419},
  shorttitle = {Mars 2020 site-specific mission performance analysis},
  title = {Mars 2020 site-specific mission performance analysis: {Part} 2. {Surface} traversability},
  year = {2018}
}

@article{papadimitriou_shortest_1991,
  author = {Papadimitriou, Christos H. and Yannakakis, Mihalis},
  doi = {10.1016/0304-3975(91)90263-2},
  issn = {0304-3975},
  journal = {Theoretical Computer Science},
  language = {en},
  month = {July},
  number = {1},
  pages = {127--150},
  title = {Shortest paths without a map},
  urldate = {2020-12-11},
  volume = {84},
  year = {1991}
}

@article{polychronopoulos_stochastic_1996,
  author = {Polychronopoulos, George H. and Tsitsiklis, John N.},
  copyright = {Copyright © 1996 John Wiley \& Sons, Inc.},
  doi = {https://doi.org/10.1002/(SICI)1097-0037(199603)27:2<133::AID-NET5>3.0.CO;2-L},
  issn = {1097-0037},
  journal = {Networks},
  language = {en},
  number = {2},
  pages = {133--143},
  title = {Stochastic shortest path problems with recourse},
  urldate = {2020-12-23},
  volume = {27},
  year = {1996}
}

@inproceedings{rankin_perseverance_2023,
  author = {Rankin, Arturo and Del Sesto, Tyler and Hwang, Pauline and Justice, Heather and Maimone, Mark and Verma, Vandi and Graser, Evan},
  booktitle = {2023 {IEEE} {Aerospace} {Conference}},
  doi = {10.1109/AERO55745.2023.10115835},
  month = {March},
  pages = {1--16},
  title = {Perseverance {Rapid} {Traverse} {Campaign}},
  urldate = {2023-10-18},
  year = {2023}
}

@inproceedings{rigter_planning_2022,
  author = {Rigter, Marc and Duckworth, Paul and Lacerda, Bruno and Hawes, Nick},
  booktitle = {Proceedings of the {International} {Conference} on {Automated} {Planning} and {Scheduling}},
  pages = {307--315},
  title = {Planning for risk-aversion and expected value in {MDPs}},
  urldate = {2023-12-01},
  volume = {32},
  year = {2022}
}

@inproceedings{rigter_risk-averse_2021,
  author = {Rigter, Marc and Lacerda, Bruno and Hawes, Nick},
  booktitle = {Advances in {Neural} {Information} {Processing} {Systems}},
  pages = {1142--1154},
  publisher = {Curran Associates, Inc.},
  title = {Risk-{Averse} {Bayes}-{Adaptive} {Reinforcement} {Learning}},
  urldate = {2024-03-13},
  volume = {34},
  year = {2021}
}

@article{rockafellar_conditional_2002,
  author = {Rockafellar, R. Tyrrell and Uryasev, Stanislav},
  doi = {10.1016/S0378-4266(02)00271-6},
  issn = {0378-4266},
  journal = {Journal of Banking \& Finance},
  month = {July},
  number = {7},
  pages = {1443--1471},
  title = {Conditional Value-at-Risk for General Loss Distributions},
  urldate = {2025-12-28},
  volume = {26},
  year = {2002}
}

@article{ruszczynski_risk-averse_2010,
  author = {Ruszczyński, Andrzej},
  doi = {10.1007/s10107-010-0393-3},
  issn = {1436-4646},
  journal = {Mathematical Programming},
  language = {en},
  month = {October},
  number = {2},
  pages = {235--261},
  title = {Risk-averse dynamic programming for {Markov} decision processes},
  urldate = {2024-03-21},
  volume = {125},
  year = {2010}
}

@inproceedings{tamar_policy_2015,
  author = {Tamar, Aviv and Glassner, Yonatan and Mannor, Shie},
  publisher = {Citeseer},
  title = {Policy gradients beyond expectations: {Conditional} value-at-risk},
  year = {2015}
}

@inproceedings{verma_first_2022,
  author = {Verma, Vandi and Hartman, Frank and Rankin, Arturo and Maimone, Mark and Del Sesto, Tyler and Toupet, Olivier and Graser, Evan and Myint, Steven and Davis, Kevin and Klein, Douglas and Koch, Justin and Brooks, Sawyer and Bailey, Philip and Justice, Heather and Dolci, Marco and Ono, Hiro},
  booktitle = {2022 {IEEE} {Aerospace} {Conference} ({AERO})},
  doi = {10.1109/AERO53065.2022.9843204},
  month = {March},
  pages = {1--20},
  title = {First 210 solar days of {Mars} 2020 {Perseverance} {Robotic} {Operations} - {Mobility}, {Robotic} {Arm}, {Sampling}, and {Helicopter}},
  year = {2022}
}

@inproceedings{wang_adaptive_2021,
  author = {Wang, Ziyi and So, Oswin and Lee, Keuntaek and Theodorou, Evangelos A.},
  booktitle = {Proceedings of the 3rd {Conference} on {Learning} for {Dynamics} and {Control}},
  editor = {Jadbabaie, Ali and Lygeros, John and Pappas, George J. and A.\&nbsp;Parrilo, Pablo and Recht, Benjamin and Tomlin, Claire J. and Zeilinger, Melanie N.},
  month = {June},
  pages = {510--522},
  publisher = {PMLR},
  series = {Proceedings of {Machine} {Learning} {Research}},
  title = {Adaptive {Risk} {Sensitive} {Model} {Predictive} {Control} with {Stochastic} {Search}},
  volume = {144},
  year = {2021}
}

@article{wang_risk-averse_2022,
  author = {Wang, Yuheng and Chapman, Margaret P.},
  doi = {10.1016/j.artint.2022.103743},
  issn = {0004-3702},
  journal = {Artificial Intelligence},
  language = {en},
  month = {October},
  pages = {103743},
  shorttitle = {Risk-averse autonomous systems},
  title = {Risk-averse autonomous systems: {A} brief history and recent developments from the perspective of optimal control},
  urldate = {2022-11-28},
  volume = {311},
  year = {2022}
}

\clearpage
\section{Appendix: Completeness and Optimality of CVaR-CTP-AO*}

In this appendix, we prove the completeness and optimality of the CVaR-CTP-AO* algorithm described in \Cref{sec:approach} and shown in \Cref{alg:cvar-aostar-main}.
We make the following assumptions at the outset.
We assume finite CVaR-CTP instances. In particular, the fully expanded AO tree contains finitely many nodes, each with finitely many successors. Transition costs are nonnegative. For every AND node $n$, the successor probabilities satisfy $\sum_{n' \in \text{succ}(n)} \text{prob}(n,n') = 1$ and $\text{prob}(n,n') \ge 0$. Terminal nodes have no successors.

We first analyze the inner optimization subproblem defined in \Cref{eq:inner_exp}.
\Cref{lem80:termination-inner} shows that the algorithm solves this  subproblem terminates after finitely many operations.
\Cref{lem80:admissibility-inner} establishes that, under an admissible heuristic, the expectations computed during the inner optimization provide admissible lower bounds on the corresponding expectations in the fully expanded tree and are exact whenever a node is marked as solved.

Next, we show that CVaR-CTP-AO* terminates after finitely many operations (\Cref{lem80:termination}).
We then establish that the shift optimization need only consider the finite set of candidate shift values induced by \edit{frontier node costs} (\Cref{lem80:shift-sufficiency}).
These results are finally used to prove completeness and optimality of CVaR-CTP-AO* (\Cref{thm80:completeness-optimality}).

\begin{lemma}[Termination of the Inner Optimization Subproblem]
\label{lem80:termination-inner}
Let $\mT$ be any finite AO search tree and let $S$ be any
finite set of shift values. During a single iteration of CVaR-CTP-AO*, the inner optimization subproblem (initializing values for newly created nodes, propagating updated values throughout $\mT$, and updating the best running partial policy for each shift value) terminates after finitely many operations.
\end{lemma}

\begin{proof}
The inner optimization block begins by initializing costs for the successors of the recently expanded node and for the shift values $s \in S$. Since the branching factor of $\mT$ is finite, only finitely many successors are processed, and a bounded amount of work is performed for each $s \in S$. Thus the initialization step terminates.

Next, the full-tree propagation step (\textsc{BackpropTree}) visits each node of $\mT$ exactly once, in order of decreasing depth. At each node, the update rule involves only its finitely many successors and a subset of shift values in $S$. Because both $\mT$ and $S$ are finite, this step also terminates.

Finally, the backpropagation procedure (\textsc{BackpropAncestors}) applies the same update rule along the chain of ancestors from the expanded node to the root of $\mT$. Since $\mT$ is finite, this chain contains only finitely many nodes, and each update again involves finitely many successors and shift values. Hence this step terminates as well.
\end{proof}

\begin{lemma}[Admissibility of the Inner Optimization Subproblem]
\label{lem80:admissibility-inner}
Let $n$ be any node of the AND-OR search tree and let $s>0$ be a fixed shift value.
Let $w(n,s)$ denote the expectation maintained by the algorithm after an iteration of the inner optimization subproblem of CVaR-CTP-AO*.
Let $h:\mX\rightarrow\mathbb{R}_{\ge0}$ be an admissible heuristic cost-to-go function used to initialize frontier node costs.
Then the values maintained by the inner optimization are admissible at every node, that is, $w(n,s)\le \hat w(n,s)$, where $\hat w(n,s)$ denotes the value obtained by applying the same AND-OR recursion to the fully expanded search tree.
Moreover, whenever a node $n$ is marked as solved for a given shift value $s$, we have $w(n,s)=\hat w(n,s)$ and the search tree encodes a complete and optimal history-dependent policy from $n$ for that shift value.
\end{lemma}

\begin{proof}
The proof follows the standard backward-induction argument used for AO* algorithms, adapted to our setting, and follows the same reasoning as \cite[Lemma~1]{martelli_additive_1973}.

Fix a shift value $s>0$. The inner optimization subproblem defines $w(n,s)$ recursively over the search tree using the same AND-OR recursion as the fully expanded value $\hat w(n,s)$: frontier nodes are initialized using the heuristic $h$, AND nodes compute probability-weighted expectations over their successors, and OR nodes select the minimum successor value.

We prove by backward induction over this recursion that after every inner optimization iteration the stored values satisfy
$w(n,s)\le\hat w(n,s)$ for all nodes $n$.

\smallskip
\noindent\textit{Base case.}
Let $n$ be a frontier node. The algorithm initializes
\begin{equation*}
w(n, s) = \bigl(g(n) + h(n) - s\bigr)^+.
\end{equation*}
Since the heuristic $h$ is admissible, $g(n) + h(n)$ does not overestimate the true expectation from $n$, and therefore
\begin{equation*}
 w(n, s) \le \hat{w}(n, s).
\end{equation*}
If $n$ is terminal, then $h(n) = 0$ and
\begin{equation*}
w(n,s) = \bigl(g(n) - s\bigr)^+ = \hat{w}(n, s),
\end{equation*}
so the value is exact and the node is marked as solved.

\smallskip
\noindent\textit{Inductive case.}
Assume $w(n',s)\le\hat w(n',s)$ holds for all successors $n'\in\mathrm{succ}(n)$.
If $n$ is an AND node,
\begin{equation*}
w(n,s)=\sum_{n'\in\mathrm{succ}(n)}\text{prob}(n,n')\,w(n',s).
\end{equation*}
Because the transition probabilities are nonnegative,
\begin{equation*}
w(n,s)\le
\sum_{n'\in\mathrm{succ}(n)}\text{prob}(n,n')\,\hat w(n',s)
=\hat w(n,s).
\end{equation*}
If $n$ is an OR node,
\begin{equation*}
w(n,s)=\min_{n'\in\mathrm{succ}(n)} w(n',s),
\end{equation*}
and therefore
\begin{equation*}
w(n,s)\le
\min_{n'\in\mathrm{succ}(n)}\hat w(n',s)
=\hat w(n,s).
\end{equation*}

Node $n$ is marked as solved for the shift value $s$ only once the relevant successor values are exact.
For an OR node, the minimizing successor $n'_{\mathrm{best}}$ is solved, so
\begin{equation*}
w(n,s)=w(n'_{\mathrm{best}},s)=\hat w(n'_{\mathrm{best}},s)=\hat w(n,s).
\end{equation*}
For an AND node, all successors are solved and exact, yielding the same equality.
When $n$ is marked as solved for a given shift value, the selected successors define a complete and optimal history-dependent policy from $n$ for the given shift value.
\end{proof}

\begin{lemma}[Termination of CVaR-CTP-AO*]
\label{lem80:termination}
Consider a finite CVaR-CTP instance with at least one complete policy and assume that all edge costs in the corresponding graph are nonnegative.
CVaR-CTP-AO* terminates after finitely many operations with the root node marked as solved.
Furthermore, the set of candidate shift values $S$ generated during the search is finite.
\end{lemma}

\begin{proof}
By assumption, the fully expanded AO search tree contains finitely many nodes, and each node has finitely many successors.
During execution of CVaR-CTP-AO*, candidate shift values are introduced only when nodes are generated, through their $f$-costs, $f(n)=g(n)+h(n)$.
Since only finitely many nodes can be generated, only finitely many distinct shift values can arise. Let $S$ denote this set.

The algorithm maintains values $w(n,s)$ only for node--shift pairs $(n,s)$ with $n$ in the search tree and $s \in S$.
Since both the number of nodes and the size of $S$ are finite, only finitely many such pairs exist.
Moreover, node--shift pairs are introduced by set union, so each pair can be created at most once.

By \Cref{lem80:termination-inner}, each inner optimization subproblem requires only finitely many operations to solve.
Hence each iteration of CVaR-CTP-AO* performs a finite amount of work and can create only node--shift pairs drawn from a finite set.
Since each such pair can be created at most once, only finitely many iterations can occur.

It remains to show that the root node is eventually marked as solved.
Since a complete policy exists and the fully expanded search tree is finite, there exists a finite complete policy subtree starting at the root node.
Because the algorithm adds nodes monotonically and only finitely many nodes can ever be generated, all nodes in this complete policy subtree are eventually generated.

For any candidate shift value $s \in S$, terminal node--shift pairs are marked as solved by the base case of the inner optimization recursion.
After each iteration, the solved information is propagated upward by the same AND-OR recursion used to update the values: an AND node is marked solved for $s$ once all of its successors are solved for $s$, and an OR node is marked solved for $s$ once its selected minimizing successor is solved for $s$.
Since the complete policy subtree is finite, repeated application of this rule \edit{(via the \textsc{BackpropTree} and \textsc{BackpropAncestors} routines)} marks its root node as solved after finitely many propagations.

Thus, after finitely many operations, the root node is marked as solved for at least one shift value and the algorithm terminates.
\end{proof}

\begin{lemma}[Sufficiency of the Set of Candidate Shift Values]
\label{lem80:shift-sufficiency}
Let $S$ denote the set of all distinct $f$-costs, $f(n)=g(n)+h(n)$, of terminal and non-terminal frontier nodes generated during the execution of CVaR-CTP-AO*. Then:
\begin{enumerate}
\itemsep=2pt
    \item The set $S$ is finite.
    \item At any iteration of CVaR-CTP-AO*, for any partial or full\footnote{\edit{A ``full'' policy is a policy in which all frontier nodes correspond to terminal states. In \Cref{sec:approach}, we called such policies ``complete'', but we adopt the term ``full'' here to avoid confusion with algorithmic notions of completeness.}} history-dependent policy $\pi$ represented by the current search tree, the discrete cost distribution induced by $\pi$ under the current tree has support contained in $S$.
    \item For any such policy $\pi$, there exists an optimal shift value $s^\star \in S$ for the minimization in \Cref{eq:opt_mod}.
    \item In particular, if the root node is marked as solved for some shift value, then the full policy encoded has an optimal shift value $s^\star \in S$.
\end{enumerate}
\end{lemma}

\begin{proof}
By \Cref{lem80:termination}, the set of candidate shift values $S$ generated during the search is finite.

Fix an iteration of CVaR-CTP-AO* and consider any history-dependent policy $\pi$ represented by the current search tree.
Under the current tree, each realization of $\pi$ terminates at a frontier node.
The corresponding realized cost is exactly the $f$-cost of that frontier node. Therefore the support of the discrete random cost induced by $\pi$ under the current tree satisfies
\begin{equation*}
\operatorname{supp}(\mathrm{C}^\pi) \subseteq S.
\end{equation*}
Recall from \Cref{sec:problemstatement} that the CVaR at risk aversion level $\alpha$ of a real-valued random cost $\mathrm{C}$ can be found by solving the convex optimization problem
\begin{equation}
\label{eq80:cvar-continuous}
\cvar_\alpha(\mathrm{C}) = \inf_{s \in \mathbb{R}} \Bigl\{ s + \frac{1}{\alpha}\mathbb{E}\bigl[(\mathrm{C}-s)^+\bigr]\Bigr\},
\end{equation}
where the minimizer is $s^*$ and corresponds to VaR$(\mathrm{C})$.
Since the cost $\mathrm{C}$ is a discrete random variable, the convex objective function in \Cref{eq80:cvar-continuous} is piecewise linear~\cite[Section 3]{rockafellar_conditional_2002} and a minimizer $s^*$ occurs at one of the breakpoints of the objective function.
These breakpoints are exactly the support values of $\mathrm{C}^\pi$.
Hence, since $\operatorname{supp}(\mathrm{C}^\pi)\subseteq S$, there exists an optimal shift value $s^\star \in S$ for the policy $\pi$.

Finally, suppose the root node is marked as solved for some shift value.
Then, by \Cref{lem80:admissibility-inner}, the marked successors encode a full history-dependent policy and the maintained value at the root is exact.
The induced cost distribution of the policy is discrete and supported on values in $S$.
Therefore its optimal shift value $s^\star$ is contained in $S$.
\end{proof}

\begin{figure*}
    \begin{subfigure}{0.47\textwidth}
        \centering
        \includegraphics[width=\textwidth]{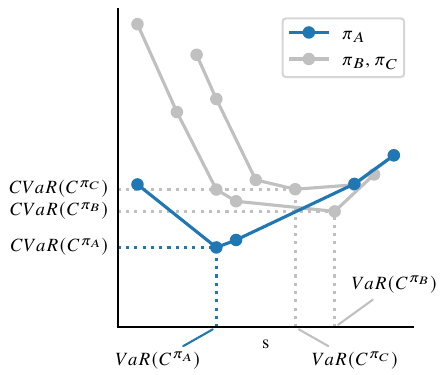}
        \caption{Three different policies with different CVaR, one of which ($\pi_A$, in blue) is lower and thus, better.}
        \label{fig80:lemma-a}
    \end{subfigure}
    \hfill
    \begin{subfigure}{0.47\textwidth}
        \centering
        \includegraphics[width=\textwidth]{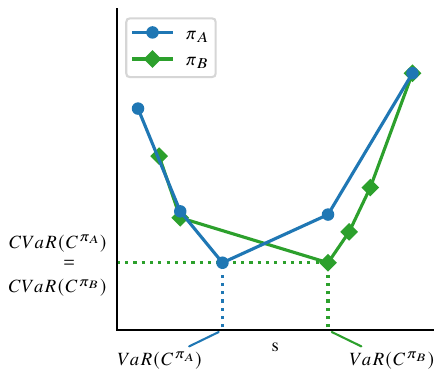}
        \caption{Two different policies with identical CVaR, but different VaR.}
        \label{fig80:lemma-b}
    \end{subfigure}
    \caption{
        Visualization of the search space of two arbitrary CVaR-CTP instances as defined in \Cref{eq:opt}.
        Each curve represents a policy, and the horizontal position of each marker represents the support of the corresponding cost distributions (and therefore the set of candidate shift values $S$).
        Each curve is the piecewise linear form of the convex objective function in \Cref{eq80:cvar-continuous} for each policy and their minimum corresponds to the CVaR of the corresponding policy.
    }
    \label{fig80:cvar-policies}
\end{figure*}

The convexity and piecewise linearity of \Cref{eq80:cvar-continuous} for discrete random variables is conceptually illustrated in \Cref{fig80:cvar-policies}.
The figure depicts the search space of CVaR-CTP instances, where each convex and piecewise linear curve corresponds to the (discrete) random cost of some partial or full history-dependent policy.
These visualizations can also be interpreted as a state of the CVaR-CTP-AO* algorithm frozen in time.
The unique breakpoint abscissas of all policies form the set of candidate shift values.
Expanding the search tree modifies the set of policies, which is equivalent to adding curves to these plots or expanding the support of existing ones.

\begin{theorem}[Completeness and Optimality of CVaR-CTP-AO*]
\label{thm80:completeness-optimality}
Consider a finite CVaR-CTP instance and assume that at least one full policy exists. Let the heuristic function $h$ be admissible.
Then CVaR-CTP-AO* terminates after finitely many operations with the root node marked as solved for at least one shift value.
Moreover, the policy encoded from the root is a full history-dependent policy that minimizes the CVaR objective. %
\end{theorem}

\begin{proof}
By \Cref{lem80:termination}, CVaR-CTP-AO* terminates after finitely many operations, with the root node marked as solved for at least one shift value, and with a finite set of candidate shift values $S$.

Let $s^\star$ be the shift value selected by the algorithm at termination, and let $\pi^\star$ denote the policy encoded by the marked successors at the root.
Since the root is marked as solved for $s^\star$,
\Cref{lem80:admissibility-inner} implies that
\begin{equation*}
w(n_{\mathrm{root}},s^\star)=\hat{w}(n_{\mathrm{root}},s^\star),
\end{equation*}
and that $\pi^\star$ is a full history-dependent policy whose value is exact for that shift.
Therefore, whenever a full policy exists, CVaR-CTP-AO* returns a full policy.

It remains to show that no other full policy can attain a smaller CVaR value.
Let $\pi$ be any full policy in the fully expanded tree, and let $\pi_A$ be the maximal partial policy represented in the current search tree that agrees with $\pi$.
Along any realization of $\pi_A$, execution terminates at a frontier node $n$, to which the current tree assigns surrogate cost $f(n)=g(n)+h(n)$.
Since $h$ is admissible, this surrogate cost does not overestimate the true total cost of any completion of $\pi_A$ through $n$.
Therefore the surrogate random cost induced by $\pi_A$ first-order stochastically dominates the true random cost induced by $\pi$, and by monotonicity of CVaR,

\begin{equation*}
\cvar_\alpha(\mathrm{C}^{\pi_A})
\le
\cvar_\alpha(\mathrm{C}^{\pi}).
\end{equation*}

\edit{If a suboptimal policy was returned by CVaR-CTP-AO*, then it would indicate that the CVaR associated with the full optimal policy tree (or any of its partial subtrees found so far) is overestimated.
An overestimation of the CVaR would imply an overestimation of frontier node costs, contradicting the assumption that heuristic function $h$ is admissible.}

Finally, by \Cref{lem80:shift-sufficiency}, the full policy $\pi^\star$ has the optimal shift value contained in the finite candidate set $S$.
Hence the outer minimization over $S$ attains an optimal shift value for $\pi^\star$.
Therefore $\pi^\star$ is complete and minimizes the CVaR objective.
\end{proof}
\end{document}